\documentclass{article}

\usepackage{amssymb}

\usepackage{graphicx}
\usepackage{tikz}
\usetikzlibrary{shapes,shadows}
\usepgflibrary{shapes}
\usepackage{pgf,pgfarrows,pgfnodes,pgfautomata,pgfheaps,pgfshade}

\usepackage{times}
\usepackage{helvet}
\usepackage{courier}
\usepackage{graphicx}
\usepackage{ifthen}
\usepackage{amsmath, amssymb}
\usepackage{eufrak}
\usepackage {float}
\usepackage[amsmath,thmmarks]{ntheorem}
\usepackage{mathrsfs}
\usepackage{verbatim}
\usepackage{array,color}

\usepackage[vlined,ruled]{algorithm2e}

\newtheorem{defin}{Definition}
\newtheorem{prop}{Proposition}
\newtheorem{coro}{Corollary}
\newtheorem{lemma}{Lemma}
\newtheorem{example}{Example}

\def\p{{\sf P} }
\def\np{{\sf NP}}
\def\npc{{\sf NP}-complete}

\def\bh2{$\mbox{{\sf BH}}_2$ }
\def\cobh2{$\mbox{{\sf coBH}}_2$ }

\def\eqbf2{{\sc qbf}$_{2,\exists}$ }

\newcommand{\ra}{\rightarrow}

\def\qed{\hfill {\scriptsize{$\blacksquare$}}\break}
\def\beginproof{\noindent {\em Proof:~}}

\def\endproof{\hfill \qed\\}

\def\iff{if and only if }

\newcommand\jl[1]{{ #1}} % blue
\newcommand\yc[1]{{ #1}}% red
\newcommand\jm[1]{{ #1}} % green
\newcommand{\nicobf}[1] {{ #1}} % magenta

\newcommand\algoname[1]{{\textit{#1}}}

\def\P{{\cal P}}

\usepackage{color}

\newcommand{\ofootnote}[1]{}

\def\P{{\cal P}}

\newcommand{\Omit} [1] {}

\title{New Candidates Welcome! \\  Possible Winners with respect to the Addition of New Candidates}

\author{
Y. Chevaleyre \\ LIPN \\ Univ. Paris 13
\and J. Lang \\ LAMSADE \\ Univ. Paris-Dauphine 
\and N. Maudet \\ LIP6 \\ Univ. Pierre et Marie Curie
\and J. Monnot \\ LAMSADE \\ Univ. Paris-Dauphine 
\and L. Xia \\ SEAS \\ Harvard University
}

\date{\today}

\begin{document}
\maketitle

\begin{abstract}
In voting contexts, some new candidates may show up in the course of the
process. In this case, we may want to determine which of the initial
candidates are possible winners, given that a fixed number $k$ of new candidates
will be added. We give a computational study of this problem, focusing on scoring
rules, and we provide a formal comparison with related problems such as control via adding candidates
or cloning.
\end{abstract}

\section{Introduction}\label{intro}
In many real-life collective decision making situations, the set of
candidates (or alternatives) may vary while the voting process goes
on, and may change at any time before the decision is final: some
new candidates may join, whereas some others may withdraw. This, of
course, does not apply to situations where the vote takes place in a
very short period of time (such as, typically, political elections
in most countries), and neither does the addition of new candidates during the
process apply to situations where the law forbids
new candidates to be introduced after the voting process has started (which,
again, is the case for most political elections). However, there are
quite many practical settings where this may happen,
especially situations where votes are sent by email during an extended
period of time. This is typically the case when making a decision
about the date and time of a meeting. In the course of the process,
we may learn that the room is taken at a given time slot, making
this time slot no longer a candidate. The opposite case also occurs
frequently; we thought the room was taken on a given date and then
we learn that it has become available, making this time slot a new
candidate.

The paper focuses on candidate addition only. More precisely, the
class of situations we consider is the following. A set of voters
have expressed their votes about a set of (initial) candidates. Then
some new candidates declare their intention to participate in the election. The winner will ultimately be
determined using some given voting rule and the voters' preferences over the set of all candidates. 
In this class of
situations, an important question arises: {\em who among the
initial candidates can still be a winner once the voters'
preferences about all candidates are known?} This is important in
particular if there is some interest to detect as soon as possible
the candidates who are not possible winners: for instance,
candidates for a job may have the opportunity to apply for different
positions, and time slots may be released for other potential
meetings.

This question is strongly related to several streams of work in
the recent literature on computational social choice, especially the
problem of determining whether the vote elicitation process can be
terminated \cite{ConitzerSandholm02b,Walsh08}; the possible winner
problem, and more generally the problem of applying a voting rule to
incomplete preferences
\cite{KonczakLang05,PiniRVW07,XiaConitzer11,BetzlerDorn09,BetzlerHemmannNiedermeier09}
or uncertain preferences with probabilistic information
\cite{HAKW09};  swap bribery, encompassing the possible winner
problem as a particular case \cite{ElkindFaliszewskiSlinko09}; 
voting with an unknown set of available candidates \cite{LuBoutilier10};
the control of a voting rule by the chair via adding
candidates; and resistance to cloning---we shall come back to
the latter two problems in more detail in the related work
section.

Clearly, considering situations where new voters are added is a
specific case of voting under incomplete preferences, where
incompleteness is of a very specific type: the set of candidates is
partitioned in two groups (the initial and the new candidates), and
the incomplete preferences consist of complete rankings on the
initial candidates. This class of situations is, in a sense, dual to a class of situations that has been considered more often, namely,
when the set of {\em voters} is partitioned in two groups: those
voters who have already voted, and those who have not expressed their
votes yet. The latter class of situations, while being a subclass of
voting under incomplete preferences, has been more specifically
studied as a {\em coalitional manipulation problem} 
\cite{ConitzerSandholmLang07,FaliszewskiProcaccia10}, where the problem is to
determine whether it is possible for the voters who have not voted
yet to make a given candidate win. Varying sets of voters have also
been studied in the context of compiling the votes of a
subelectorate \cite{CLMR09,XiaConitzerAAAI10}: there, one is
interested in summarizing a set of initial votes, while still being
able to compute the outcome once the remaining voters have expressed
their votes.

The layout of the paper is as follows. In Section \ref{background}
we recall the necessary background on voting and we introduce some
notation. In Section \ref{possible} we state the problem formally,
by defining voting situations where candidates may be added after
the votes over a subset of initial candidates have already been
elicited. In the following sections we focus on specific voting
rules and we study the problem from a computational point of view.
In Section \ref{kapp}, we focus on the family of $K$-approval rules,
including plurality and veto as specific subcases, and give a full
dichotomy result for the complexity of the possible winner problem with
respect to the addition of $k$ new candidates; namely, we show that
the problem is {\sf NP}-complete as soon as $K \geq 3$ and $k \geq
3$, and polynomial if $K \leq 2$ or $k \leq 2$. In Section
\ref{borda} we focus on the Borda rule and show that the problem is
polynomial-time solvable regardless of the number of new candidates. We also exhibit a more general family of voting rules, including Borda, for which this result can be generalized. In Section
\ref{hardone} we show that the problem can be hard for some
positional scoring rules even if only one new candidate is added. In
Section \ref{related} we discuss the relationship to the general
possible winner problem, to the control of an election by the chair
via adding candidates, and to candidate cloning. Section
\ref{conclu} summarizes the results and mentions further research
directions.

\section{Background and notation}\label{background}

Let $C$ be a finite set of {\em candidates}, and $N$ %=\{v_1,\ldots,v_n\}$ be
a finite set of {\em voters}. The number of voters is denoted by $n$, and the (total) number of candidates
by $m$. 
A {\em $C$-vote} (called simply a vote when this is not ambiguous) is a linear order over $C$, denoted by $\succ$ or by $V$.
We sometimes denote votes in the following way:  $a \succ b \succ c$ is denoted by $abc$, etc.
An $n$-voter
{\em $C$-profile} %when there is no ambiguity about the number of voters,
is a collection $P = \langle V_1,\ldots, V_n\rangle$ of $C$-votes.
Let $\P_C$ be the set of all $C$-votes and therefore $\P_C^n$ be the set of all $n$-voter $C$-profiles.
We denote by $\P_C^*$ the set of all $n$-voter $C$-profiles for $n \geq 1$, {\em i.e.}, $\P_C^* = \cup_{n \geq 1} \P_C^n$.

A voting rule on $C$ is a function $r$ from $\P_C^*$ to $C$.
A voting correspondence is a function from $\P_C^*$ to $2^C\setminus\{\emptyset\}$.
The most natural way of obtaining a voting rule from a voting correspondence is to break
ties according to a fixed priority order on candidates. In this paper, we
do not fix a priority order on candidates (one reason being that the
complete set of candidates is not known to start with), which means
that we consider voting correspondences rather than rules,
and ask whether $x$ is a {\em possible cowinner} for a given profile
$P$. This is equivalent to asking whether there exists a priority
order for which $x$ is a possible winner, or else whether $x$ is a
possible winner for the most favorable priority order (with $x$
having priority over all other candidates).
This is justified in our context by the fact that specifying such a priority order is problematic when we don't know in advance the identities of the potential new candidates. 
With a slight abuse of notation we denote voting correspondences by $r$ just as voting rules. Let $r(P)$ be the set of {\em cowinners} for profile $P$.

For $P \in \P_C^*$ and $x,x' \in C$, let $n(P,i,x)$ be the number of votes in $P$ ranking $x$ in position $i$,
$ntop(P,x) = n(P,1,x)$ the number of votes in $P$ ranking $x$ first, and $N_P(x,x')$ the number of votes in $P$ ranking $x$ above $x'$.
Let $\vec{s} = \langle s_1, \ldots, s_m\rangle$ be a vector of integers such that $s_1 \geq \ldots \geq s_m$ and $s_1 > s_m$.
The scoring rule  $r_{\vec{s}}(P)$ induced by $\vec{s}$ elects the candidate(s) maximizing
$S_{\vec{s}}(x,P) = \sum_{i=1}^m s_i \cdot n(P,i,x)$. 

If $K$ is a fixed integer then  {\em $K$-approval}, $r_K$, is the scoring rule corresponding to the vector $\vec{s_K} = \langle 1,\ldots, 1, 0, \ldots, 0\rangle$ -- with $K$ 1's and $m-K$ 0's.  The $K$-approval score $S_{\vec{s_K}}(x,P)$ of a candidate $x$ is denoted more simply by $S_K(x,P)$: in other words, $S_K(x,P)$ is the number of voters in $P$ who rank $x$ in the  first $K$ positions, {\em i.e.}, $S_K(x,P) = \sum_{i = 1, \ldots, K} n(P,i,x)$.  When $K = 1$, we get the {\em plurality} rule $r_P$, and when $K = m-1$ we get the {\em veto} (or {\em antiplurality}) rule.
The {\em Borda} rule $r_B$ is the scoring rule corresponding to the vector $\langle m-1,m-2,\ldots,0\rangle$. 

We now define formally situations where new candidates are added.

\begin{defin}\label{situation}
A {\em voting situation with a varying set of candidates} is a 4-tuple $\Sigma = \langle N, X, P_X, k \rangle$ where $N$ is a set of voters (with $|N| = n$), $X$ a set of candidates, $P_X = \langle V_1,\ldots, V_n \rangle$ an $n$-voter $X$-profile, and  $k$ is a positive integer, encoded in unary.
\end{defin}

$X$ denotes the set of initial candidates, $P_X$ the initial profile, and $k$ the number of new candidates. 
Nothing is known a priori about the voters' preferences over the new candidates, henceforth their identity is irrelevant and only their number counts. The assumption that $k$ is encoded in unary ensures that the number of new candidates is polynomial in the size of the input. Most of our results would still hold if the number of new candidates is exponentially large in the size of the input, but for the sake of simplicity, and also because, in practice, $k$ will be small anyway, we prefer to exclude this possibility.

Because the number of candidates is not the same before and after the new candidates come in, we have to consider families of voting rules (for a varying number of candidates) rather than voting rules for a fixed number of candidates.
While it is true that for many usual voting rules there is an obvious way of defining them for a varying number of candidates, this is not the case for all of them, especially scoring rules. Still, some natural scoring rules, including plurality, veto, more generally $K$-approval, as well as Borda, are naturally defined for any number of candidates.  We shall therefore consider {\em families of voting rules}, parameterized by the number of candidates ($r^m$). We slightly abuse notation and denote these families of voting rules by $r$, and consequently %usually
often write $r(P)$ instead of $r^m(P)$. The complexity results we give in this paper make use of such families of voting rules, where the number of candidates is variable. 

If $P$ is a $C$-profile and $C' \subseteq C$, then the projection of $P$ on $C'$, denoted by $P^{\downarrow C'}$, is obtained
by deleting all candidates in $C \setminus C'$ in each of the votes of $P$, and leaving unchanged the ranking on the candidates of $C'$.
For instance, if $P = \langle abcd, dcab \rangle$, then $P^{\downarrow \{a,b\}} = \langle ab, ab\rangle$ and
 $P^{\downarrow \{a,b,c\} }= \langle abc, cab\rangle$.
In all situations, the set of initial candidates is denoted by $X =
\{x_1,\ldots, x_p\}\cup \{x^*\}$, the set of the $k$ new candidates
is denoted by $Y = \{y_1, \ldots, y_k\}$. If $P_X$ is an $X$-profile
and $P'$ an $X \cup Y$-profile, then we say that $P'$ extends $P_X$
if the projection of $P'$ on $X$ is exactly $P_X$.
For instance, let $X = \{x_1,x_2,x_3\}\cup \{x^*\}$, $Y =
\{y_1,y_2\}$; the profile $P'= \langle x_1 y_1 x^* x_2 y_2 x_3, y_1
y_2 x_1 x_2 x_3 x^*, x_3 x_2 y_2 x^* y_1  x_1\rangle$ extends the
$X$-profile $P_X= \langle x_1 x^* x_2 x_3, x_1 x_2 x_3 x^*, x_3 x_2
x^* x_1\rangle$.

\section{Possible winners when new candidates are added}\label{possible}

We recall from \cite{KonczakLang05} that given a collection $\langle
P_1, \ldots, P_n\rangle$ of partial strict orders on $C$
representing some incomplete information about the votes, a
candidate $x^*$ is a possible winner if there is a profile  $\langle
T_1, \ldots, T_n\rangle$ where each $T_i$ is a ranking on $C$
extending $P_i$ in which $x^*$ wins. Reformulated for the case where
$P_i$ is a ranking of the initial candidates (those in $X$), we get
the following definition:

\begin{defin}\label{posswin}%[possible winners with respect to the addition of candidates]
Given a voting situation  $\Sigma = \langle N, X, P_X, k \rangle$, 
and a collection $r$ of voting rules, we say that $x^* \in X$ is a {\em possible cowinner with respect to $\Sigma$ and $r$} if there is a $(X \cup Y)$-profile $P'$ extending $P_X$ such that $x^* \in r(P')$, where $Y = \{ y_1, \ldots, y_k \}$ is a set of $k$ new candidates.
\end{defin}

Note that we do not have $Y$ in the input, because it would be redundant with $k$: it is enough to know the {\em number} of new candidates. Note also that all new candidates $\{y_1,\ldots, y_k\}$ have to appear in the extended votes composing $P'$.

Also, we do not consider the problem of deciding whether a new candidate $y_j$ is a possible cowinner, because it is trivial. Indeed, as soon as  the voting correspondence  
satisfies the extremely weak property that a candidate ranked first by all voters is always a cowinner (which is obviously satisfied by all common voting rules),  any new candidate is a possible cowinner. 

We now define formally the problems we study in this paper.

\begin{defin}
Given a collection $r$ of voting rules, the {\sc possible cowinner problem with new candidates} (or {\sc PcWNC}) for $r$ is defined as follows:
\begin{description}
\item[Input] A voting situation $\Sigma = \langle N, X, P, k\rangle$ and a candidate $x^*\in X$.
\item[Question] Is $x^*$ a possible cowinner with respect to $\Sigma$ and $r$?
\end{description}
Also, the subproblem of {\sc PcWNC} where the number $k$ of new candidates is fixed will be denoted by {\sc PcWNC}$(k)$.
\end{defin}

We can also define the notion of {\em necessary cowinner} with
respect to $\Sigma$ and $r$: $x^* \in X$ is a {\em necessary
cowinner with respect to $\Sigma$, $Y$, and $r$} if for every $(X
\cup Y)$-profile $P'$ extending $P_X$ we have $x^* \in r(P)$.
However, the study of necessary cowinners in this particular setting
will almost never lead to any significant results. 
There may be necessary cowinners among the
initial candidates, but this will happen rarely (and this case
will be discussed for a few specific voting rules in the
corresponding parts of the paper).

Now we are in position to consider specific voting rules. 

\section{$K$-approval}\label{kapp}

As a warm-up we start by considering the plurality 
rule. 

\subsection{Plurality}\label{pluve}

Let us start with an example: suppose $X = \{a,b,c\}$, $n = 13$, and the plurality scores in $P_X$ are $a \mapsto 6$, $b \mapsto 4$, $c \mapsto 3$.
There is only one new candidate ($y$). We have:

\begin{enumerate}
\item
$a$ is a possible cowinner ($a$ will win in particular if the top candidate of every voter remains the same);
\item $b$  is a possible cowinner: to see this, suppose that 2 voters who had ranked $a$ first now rank $y$ first; the new scores are  $a \mapsto 4$, $b \mapsto 4$, $c \mapsto 3$, $y \mapsto 2$;
\item $c$ is not a possible cowinner: to reduce the scores of $a$ (resp. $b$) to that of $c$, we need at least 3 (resp. 1) voters who had ranked $a$ (resp. $b$) first to now rank $y$ first; but this then means that $y$ gets at least 4 votes, while $c$ has only 3.
\end{enumerate}

More generally, we have the following result:

\begin{prop}\label{prop-plura}
Let $P_X$ be an $n$-voter profile on $X$, and $x^* \in X$. The candidate $x^*$ is a possible cowinner for $P_X$ and plurality
with respect to the addition of $k$ new candidates  
\iff
$$ntop(P_X,x^*) \geq \frac{1}{k} \cdot \sum_{x_i \in X}\max(0, ntop(P_X,x_i)-ntop(P_X,x^*))$$
\end{prop}

\beginproof
Suppose first that the inequality holds.
We build the following $(X \cup Y)$-profile $P'$ extending $P_X$:
\begin{enumerate}
\item for every candidate $x_i$ such that $ntop(P_X,x_i) > ntop(P_X,x^*)$ we simply take $ntop(P_X,x_i) - ntop(P_X,x^*)$
arbitrary votes ranking $x_i$ on top and place one of the $y_j$'s on top of the vote (and the other $y_j$'s anywhere),
subject to the condition that no $y_j$ is placed on top of a vote more than $ntop(P_X,x^*)$ times. (This is possible because the inequality is satisfied).
\item in all other votes (those not considered at step 1), place all $y_j$'s anywhere except on top.
\end{enumerate}
We obtain a profile $P'$ extending $P_X$. First, we have $ntop(P', x^*) = ntop(P_X, x^*)$,
because in all the votes in $P_X$ where $x^*$ is on top, the new top candidate in the corresponding vote in $P'$ is still $x^*$
(cf. step 2), and all the votes in $P_X$ where $x^*$ was not on top obviously cannot have $x^*$ on top in the corresponding vote in $P'$.
Second, let $x_i \neq x^*$. If $ntop(P_X,x_i) \leq ntop(P_X,x^*)$ then $ntop(P',x_i) = ntop(P_X,x_i)$; and
if $ntop(P_X,x_i) > ntop(P_X,x^*)$ then we have $ntop(P',x_i) = ntop(P_X,x_i) - (ntop(P_X,x_i) - ntop(P_X,x^*)) = ntop(P_X,x^*)$.
Therefore, \jl{$x^*$ is a cowinner for plurality in $P'$}.\\
% (assuming the most favourable tie-breaking priority for $x$)
% is $x^*$.\\
Conversely, if the inequality is not satisfied, in order for $x^*$ to become \jl{a cowinner} in $P'$,
the other $x_i$'s must lose globally an amount of $\sum_{x_i \in X}\max(0, ntop(P_X,x_i)-ntop(P_X,x^*))$ votes. 
But since we have $\sum_{x_i \in X}\max(0, ntop(P_X,x_i)-ntop(P_X,x^*)) > k \cdot ntop(P_X,x^*)$, for at least one of the $y_j$'s
it must hold that $ntop(P',y_j) > ntop(P',x^*)$; therefore $x^*$ cannot be \jl{a cowinner} for plurality in $P'$.
\endproof

We do not need to pay much attention to the veto rule, since the characterization of possible cowinners is 
trivial. Indeed, by placing any of the new candidates below $x^*$ in every vote of $P_X$ where $x^*$ is ranked at the bottom position, we obtain a vote $P'$ where no one vetoes $x^*$, so any candidate is a possible cowinner.

As a corollary, computing possible cowinners for the rules of plurality (and veto) with respect to candidate addition can be computed in polynomial time (which we already knew, since possible cowinners for plurality and veto can be computed in polynomial time \cite{BetzlerDorn09}).

\subsection{$K$-approval, one new candidate}\label{kapp-1}
We start with the 
case where a single candidate is added. 
Recall that we denote by $S_K(x_j,P_X)$ the score of $x_j$ for $P_X$
and $K$-approval (\emph{i.e.} the number of voters who rank $x_j$
among their top $K$ candidates); and by $n(P_X,K,x_j)$ the number of
voters who rank $x_j$ exactly in position $K$. 

\begin{prop}\label{prop-kapp}
Let $K$ be an positive integer,  
$P_X$ be an $n$-voter profile on $X$, and $x^* \in X$. The candidate $x^*$ is a possible cowinner for $P_X$
and $K$-approval with respect to the addition of one new candidate  
\iff the following two conditions hold:
\begin{enumerate}
\item for every $x_i \neq x^*$, if $S_K(x_i,P_X) > S_K(x^*,P_X)$ \\
then $n(P_X,K,x_i) \geq S_K(x_i,P_X) - S_K(x^*,P_X)$.
\item {\small
 $S_K(x^*,P_X) \geq \sum_{x_i \in X}\max(0, S_K(x_i,P_X) - S_K(x^*,P_X))$}
\end{enumerate}
\end{prop}

\beginproof
Assume conditions (1) and (2) are satisfied. Then, we build the following $(X \cup \{y\})$-profile extending $P_X$:
\begin{itemize}
\item[(i)] for every $x_i$ such that $S_K(x_i,P_X) > S_K(x^*,P_X)$, we take $S_K(x_i,P_X) - S_K(x^*,P_X)$
arbitrary votes who rank $x_i$ in position $K$ in $P_X$ and place
$y$ on top (condition (1) ensures that we can find enough such votes).
\item[(ii)] in all other votes (those not considered at step (i)), place $y$ in the bottom position.
\end{itemize}

We obtain a profile $P'$ extending $P_X$. First, we have
$S_K(x^*,P') = S_K(x^*,P_X)$, because (a) all votes in $P_X$ ranking
$x^*$ in position $K$ are extended in such a way that $y$ is placed
in the bottom position, therefore $x^*$ gets a point in each of
these votes if and only if it got a point in $P_X$, and (b) in all the
other votes (those where $x^*$ is not ranked in position $K$ in
$P_X$), $x^*$ certainly gets a point  in $P'$ if and only if they
got a point in $P_X$. This holds both in the case where $y$ was added at the
top or the bottom of the vote. Second, for every $x_i$ such that
$S_K(x_i,P_X) > S_K(x^*,P_X)$, $x_i$ loses exactly $S_K(x_i,P_X) -
S_K(x^*,P_X)$ points when $P_X$ is extended into $P'$, therefore
$S_K(x_i,P') = S_K(x_i,P_X) - S_K(x_i,P_X) + S_K(x^*,P_X) =
S_K(x^*,P_X)$. Third, $S_K(y,P') = \sum_{x_i \in X}\max(0,
S_K(x_i,P_X) - S_K(x^*,P_X)) \leq S_K(x^*,P_X)$---because of
(2)---hence $S_K(y,P') \leq S_K(x^*,P')$. Therefore, $x^*$ is a
cowinner for $K$-approval in $P'$.

Now, assume condition (1) is not satisfied, that is, there is an
$x_i$ such that $S_K(x_i,P_X) > S_K(x^*,P_X)$ and such that $n(P_X,
K,x_i) < S_K(x_i,P_X) - S_K(x^*,P_X)$. There is no way of having
$x_i$ lose more than $S_K(x_i,P_X)$ points, therefore $x^*$ will
never catch up with $x_i$'s advantage and is therefore not a possible
\jl{cowinner}. Finally,  assume condition (2) is not satisfied, which means
that we have $\sum_{x_i \in X}\max(0, S_K(x_i,P_X) - S_K(x^*,P_X)) >
S_K(x^*,P_X)$. Then, in order for $x^*$ to reach the score of
$x_i$'s we must add $y$ in one of the top $K$ positions in a number
of votes exceeding $S_K(x^*,P_X)$, therefore $S_K(y,P')
> S_K(x^*,P_X) \geq S_K(x^*,P')$, and therefore $x^*$ is not a
possible cowinner.
\endproof

Therefore, computing possible cowinners for $K$-approval with respect to the addition of {\em one} candidate can be done in polynomial time.

%%___1
\subsection{$2$-approval, any (fixed) number of new candidates}\label{2app}

For each profile $P$ and each candidate $x'$, we simply write $s(x',P)$ for the
score of $x'$ in $P$ under $r_2$, that is,
$s(x',P)=S_2(x',P)$, \emph{i.e.} the number of times that $x'$ is
ranked within the top two positions in $P$.

Let $P_X = \langle V_1, \ldots, V_n\rangle$ be an initial profile and $Y = \{y_1,\ldots, y_k\}$ the set of new candidates. Let $x^* \in X$. We want to know whether $x^*$ is a possible cowinner for 2-approval and $P_X$. 
Let us partition $P_X$ into $P_1$, $P_2$ and $P_3$, where $P_1$
consists of the votes in which $x^*$ is ranked in the top position, $P_2$ consists of the votes in which $x^*$ is ranked in the second position and $P_3$ consists of the votes in
which $x^*$ is {\em not} ranked within the top two positions.
Let $P$ be an extension of $P_X$ to $X \cup Y$. 
For each candidate $x'\in X$, we define the following three subsets of $P$:
\begin{itemize}
\item $\text{HP}(P,x')$ is the set of votes in $P$ where $x'$ is ranked in the second
position and neither $x^*$ nor any new candidate is ranked in the top position (HP stands for
``high priority'').
\item $\text{MP}(P,x')$ is the set of votes in $P$ where $x^*$ or any new candidate is ranked in the top position and $x'$ is ranked in the second position (MP stands for ``medium priority'').
\item $\text{LP}(P,x')$ is the set of votes in $P$ where $x'$ is ranked in the top position and some $x'' \in X\setminus\{x^*\}$ is ranked in the second position (LP stands for ``low priority'').
\end{itemize}
These definitions also apply to $P_X$; our definitions then simplify into: $\text{HP}(P_X,x')$ is the set of votes in $P_X$ where $x'$ is ranked second and $x^*$ is not ranked first; $\text{MP}(P_X,x')$ is the set of votes in $P_X$ where $x^*$ is ranked first and $x'$ is ranked second; $\text{LP}(P_X,x')$ is the set of votes in $P_X$ where $x'$ is ranked first and $x^*$ is not ranked second.
These definitions are summarized in Figure \ref{HPMPLP}. 
Finally, for $x\in X\cup Y$, let $\Delta(P,x)=S_2(x,P)-S_2(x^*,P)$. 

\begin{figure}[t]
\begin{center}
\begin{tabular}{|c|c|c|}
\hline
 & top candidate belongs to & $2^{nd}$ candidate belongs to \tabularnewline
\hline
$\text{HP}(P,x')$ & $X\backslash\{x^{*}\}$ & $\{x'\}$\tabularnewline
\hline
$\text{MP}(P,x')$ & $Y\cup\{x^{*}\}$ & $\text{\{}x'\}$\tabularnewline
\hline
$\text{LP}(P,x')$ & $\{x'\}$ & $X\backslash\{x^{*}\}$\tabularnewline
\hline
\end{tabular}
\end{center}
\caption{A vote $V\in P$ belongs respectively to the sets $\text{HP}(.),\text{MP}(.),\text{LP}(.)$ if its top two candidates belong to the respective sets.
} \label{HPMPLP}
\end{figure}

Let us compute these sets on a concrete example, which will be reused throughout the section.

\begin{example}
Let $X = \{x^*, x_1, \ldots, x_6\}$ and consider the following profile $P_X$ consisting of 19 votes (we only mention the first two candidates in each vote): \\
\begin{footnotesize}
\[
\begin{array}{ccccccccccccccccccc}
v_1 & v_2 & v_3 & v_4 & v_5 & v_6 & v_7 & v_8 & v_9 & v_{10} & v_{11} & v_{12} & v_{13} & v_{14} & v_{15} & v_{16} & v_{17} & v_{18} & v_{19} \\
\hline
x^* & x_1 & x_2 & x_3 & x_1 & x_1 & x_1 & x_2 & x_2 & x_2 & x_2 & x_2 & x_3 & x_3 & x_3 & x_3 & x_3 & x_3 & x_4 \\
x_1 & x^* & x^* & x^* & x_4 & x_4 & x_5 & x_1 & x_3 & x_4 & x_5 & x_5 & x_1 & x_2 & x_4 & x_4 & x_5 & x_6& x_6 \\
\vdots & \vdots &\vdots &\vdots &\vdots &\vdots &\vdots &\vdots &\vdots &\vdots &\vdots &\vdots &\vdots &\vdots &\vdots &\vdots &\vdots &\vdots &\vdots
\end{array}
\]
\end{footnotesize}

\Omit{
\begin{tabular}{lllllll}
$v_1: c x_1$ & $v_2: x_1 c$& $v_3: x_2 c$& $v_4: x_3 c$& $v_5: x_1 x_4$& $v_6: x_1 x_4$& $v_7: x_1 x_5$ \\
$v_8: x_2 x_1$& $v_9: x_2 x_3$& $v_{10}: x_2 x_4$& $v_{11}: x_2 x_5$& $v_{12}: x_2 x_5$& $v_{13}: x_3 x_1$\\$v_{14}: x_3 x_2$& $v_{15}: x_3 x_4$& $v_{16}: x_3 x_4$& $v_{17}: x_3 x_5$& $v_{18}: x_3 x_6$& $v_{19}: x_4 x_6$
\end{tabular}\\
}

We have $P_1 = \{v_1\}$, $P_2 = \{v_2, v_3, v_4\}$ and $P_3 = \{v_5, \ldots, v_{19} \}$.
This is summarized together with the priority classification in the following table: 
$$\begin{array}{|l|l|l|l|c|}
\hline
& \mbox{{\rm HP}} & \mbox{\rm MP} & \mbox{\rm LP} & \Delta(P_X,x_i) \\ \hline
x_1 & v_8, v_{13} & v_1 & v_5, v_6, v_7 & 3\\
x_2 & v_{14} && v_8, v_9, v_{10}, v_{11}, v_{12} & 3\\
x_3 & v_9 && v_{13}, v_{14}, v_{15}, v_{16}, v_{17}, v_{18} & 4\\
x_4 & v_5, v_6, v_{10}, v_{15}, v_{16} && v_{19} & 2 \\ 
x_5 & v_7, v_{11}, v_{12}, v_{17} &  &  & 0\\
x_6 & v_{18}, v_{19} &  &  & -2\\ \hline
\end{array}$$
\end{example}

%%%%%%%%%% HERE
%

If $P^*$ is an extension of $P_X$ to $X \cup Y$ then we write 
$P^* = \langle V_1^*, \ldots, V_n^*\rangle$, where $V_i^*$ is the vote over $X \cup Y$
extending $V_i$. 
We now establish a useful property of the extensions
of $P_X$ for which $x^*$ is a cowinner. 
Without loss of generality, 
we assume that in every vote $V_i^*$, every new candidate $y_j$ is ranked either in the first two positions, or below all candidates of $X$.

\begin{prop}
\label{prop:opt} If there exists an extension $P$ of $P_X$ such that
$x^*\in r_2(P)$, then there exists an extension $P^*$ of $P_X$ such that $x^*\in r_2(P^*)$, and satisfying 
the following conditions: 
\begin{enumerate}
\item For each $V_i \in P_X$, if $x^*$ is ranked within the top two positions in $V_i$, then $x^*$ is also ranked within the top two positions in $V_i^*$.
\item For each $V^*_i\in P^*$, if the top candidate of $V_i^*$ is not in $Y$ then the second-ranked candidate of $V^*_i$ is not in $Y$ either.   
\item For each $x'\in X\setminus\{x^*\}$ and each $V_i\in \text{\rm MP}(P_X,x')\cup \text{\rm LP}(P_X,x')$, if $x'$ is not ranked within the top two positions in $V^*_i$, 
then for each $V_j\in \text{\rm HP}(P_X,x')$, $x'$ is not ranked within the top two positions in $V_j^*$.
\end{enumerate}
\end{prop}
\beginproof%[p]{prop:opt}
We consider in turn the different conditions: 
\begin{enumerate}
\item  This is because if there exists $V'\in P$ 
such that $x^*$ is not in the
top two positions whereas $x^*$ is  in the
top two positions in its original vote $V\in P_X$, then we can simply move all of candidates in
$Y$ ranked higher than $x^*$ to the bottom positions. Let
$V^*$ denote the vote obtained this way. By replacing $V'$ with
$V^*$, we increase the score of $x^*$ by $1$, and the score of each
other candidate by no more than $1$, which means that
$x^*$ is still a cowinner. 
\item If there exists $V'\in P$ such that $x'\in X$ is ranked in the
top position and $y\in Y$ is ranked in the second position, then
we simply obtain $V^*$ by switching $y$ and $x'$. 
\item The condition states that for each candidate $x'$, whenever we want to reduce
its score, we should first try to reduce it by putting a new
candidate $y\in Y$ on top of some vote in $V\in \text{HP}(P_X,x')$.
This is because by putting $y$ on top of some vote in
$\text{HP}(P_X,x')$, we may use only one extra candidate $y' \in Y$
to reduce by one unit the score of the candidate ranked at the top
position of $V$. Formally, suppose there exist $V_1\in
\text{HP}(P_X,x')$ and $V_2\in \text{MP}(P_X,x')\cup
\text{LP}(P_X,x')$ such that $x'$ is within the top two positions of
$V_1'$ (the extension of $V_1$) but not within the top two positions
of $V_2'$ (the extension of $V_2$). Let $y\in Y$ be any candidate
ranked within the top two positions of $V_2'$. Let $V_2^*$ denote
the vote obtained from $V_2'$ by moving $y$ to the bottom, and let
$V_1^*$ denote the vote obtained from $V_1'$ by moving $y$ to the
top position. Next, we replace $V_1'$ and $V_2'$ by $V_1^*$ and
$V_2^*$, respectively. It follows that the score of each candidate
does not change, which means that $x^*$ is still a cowinner. 
We repeat this procedure until statement (3) is
satisfied for every $x'\in X\setminus\{x^*\}$. Since after each
iteration there is at least one additional vote that will never be
modified again, this procedure ends in $O(|P_X|)$ times.
\end{enumerate}
\endproof

Proposition~\ref{prop:opt} simply tells us that when looking for an
extension that makes $x^*$ a cowinner, it suffices to restrict our
attention to the extensions that satisfy conditions (1) to (3). 
Moreover, using (1) of Proposition~\ref{prop:opt}, we
deduce that $s(x^*,P^*)=s(x^*,P_X)$. Hence, for votes $V\in P_2$
(the votes in which $x^*$ is ranked in the second position), we can
assume that the new candidates of $Y$ are put in bottom positions
in $P^*$. 

Define $X^\bullet$ as the set of all candidates in $X$ such that $\Delta(P_X,x_i) > 0$. 
Our objective is to reduce all score differences to $0$ for $x\in X^\bullet$, while keeping the score differences
of each new candidate non-positive. (We do not have to care about the candidates in $X \setminus X^\bullet$).

The intuition underlying our algorithm is that 
 when trying to reduce $\Delta(P,x_i)$ on the current profile $P$, we first try to use the votes in $\text{HP}(P_X,x_i)$, then the votes in $\text{MP}(P_X,x_i)$, and
finally the votes in $\text{LP}(P_X,x_i)$. This is because putting some
candidates from $Y$ in the top positions in the votes of
$\text{HP}(P_X,x_i)$ not only reduces $\Delta(P,x_i)$ by one unit, but also
creates an opportunity to 
``pay'' one extra candidate from
$Y$ to reduce $\Delta(P,x_j)$ by one unit, where $x_j$ is the candidate ranked on
top of this vote. For the votes in $\text{MP}(P_X,x_i)$, we can only
reduce $\Delta(P_X,x_i)$ by one unit without any other benefit. For the votes in
$\text{LP}(P_X,x_i)$ we will have to use two candidates from $Y$ to
bring down $\Delta(P,x_i)$ by one unit; however, if we already put some $y\in Y$ in
the top position in order to reduce $\Delta(P,x_j)$, where $x_j$ is the
candidate ranked in the second position in the original
vote, then we only need to pay one extra candidate in $Y$ to
reduce $\Delta(P,x_i)$ by one unit.
Therefore, the major issue consists in finding the most
efficient way to choose the votes in $\text{HP}(P_X,x_i)$ to reduce
$\Delta(P,x_i)$, when $\Delta(P,x_i)\leq |\text{HP}(P,x_i)|$. We will solve this problem by reducing it
to a max-flow problem.

 The algorithm is composed of a main function \algoname{CheckCowinner(.)} which comes together with two sub-functions \algoname{AddNewAlternativeOnTop(.)} and  \algoname{BuildMaxFlowGraph(.)} that we detail first. 

\LinesNumbered
\begin{algorithm}[h]
\caption{AddNewAlternativeOnTop$(P,V,Y)$} \label{algo:addnew}
$y_i\leftarrow argmin_j \left\{ \Delta(P,y_j) : y_j\in Y \right\}$ // take lowest index $i$ when tie-breaking \\
add $y_i$ on top of $V$ and update $P$\\
return $P$\\
\end{algorithm}

The procedure \algoname{AddNewAlternativeOnTop} simply picks new candidates to be put on top of votes, and updates subsequently the profile. 
Note that in this procedure, candidates from $Y$ to be added on top of the votes are those with the lowest score (or the lowest index, in case of ties). This results in choosing new candidates in a cyclic order $y_1\ra y_2\ldots\ra y_{|Y|}\ra y_1 \ldots$

As for the function \algoname{BuildMaxFlowGraph}$(P,x^{*},X_1,X_2)$, \nicobf{it} builds the weighted directed graph $G=\langle W,E \rangle$ defined as follows:
\begin{itemize}
\item $W = \{s,t\}\cup X_1\cup X_2\cup \bigcup_{x_i\in X_2} \text{LP}(P,x_i)$; \label{BMFG_startinit}
\item $E$ contains the following weighted edges:
\begin{itemize}
\item for each $x \in X_1$, an edge $(s,x)$ with weight  \jm{$\Delta(P,x)$};
\item for each $x \in X_2$ and each $V\in \text{LP}(P,x)$: an edge $(V,x)$ with weight $1$; plus, if the candidate $x'$ in second position in $V$ is in $X_1$, an edge $(x',V)$  \jm{with weight $1$};
\item  for each $x \in X_2$, an edge $(x,t)$ with weight $\Delta(P,x)$.
\end{itemize} 
\end{itemize}
We refer the reader to Figure \ref{figureflow} for an illustration. (Once this graph is constructed, any standard function to compute a flow $\phi$ of maximal value can of course be used). 
We are now in a position to detail the main function \algoname{CheckCowinner(.)}.

\LinesNumbered
\begin{algorithm}[H]
 \label{algocc}
 \caption{CheckCowinner$(P_X,x^{*},Y)$} 
 $P\leftarrow P_{X}$ \label{CC_init} \\
 $T \leftarrow 0$ // number of calls AddNewAlternativeOnTop \\
 $X_{1}\leftarrow\left\{ x_{i} \in X^\bullet:\left|\text{HP}(P_X,x_{i})\right|>\Delta(P_X,x_i)\right\} $ \\
 $X_{2}\leftarrow\left\{ x_{i} \in X^\bullet:\left|\text{HP}(P_X,x_{i})\right|\le \Delta(P_X,x_i)\right\} $\label{CC_endinit} \\
$REM \leftarrow \emptyset$ \\
\For{$x_{i}\in X_{2}$ \label{CC_for1}}{
\For{$V\in \text{HP}(P,x_{i})$}{
$P\leftarrow \algoname{AddNewAlternativeOnTop}(P,V,Y)$\\
$T\!+\!+$ \label{CC_endfor1}
}
}
\For{$x_{i}\in X_{2}$ \label{CC_for2}}{
\For{$V\in \text{MP}(P,x_{i})$ \label{CC_for22}}{
\If{$\Delta(P,x_i) > 0$  \label{CC_if1}}{
$P\leftarrow \algoname{AddNewAlternativeOnTop}(P,V,Y)$\\
$T\!+\!+$
}
\Else{$REM \leftarrow REM \cup \{x_i\}$ \label{CC_donewithx2} \label{CC_endfor2}
}
}
}
$X_2 \leftarrow X_2 \backslash REM$ \\
\If{$\exists y\in Y$ \text{such that} $\Delta(P,y)>0$  \label{CC_if2}}
{return {\em false}  \label{CC_existsy}} 
$G \leftarrow \algoname{BuildMaxFlowGraph}(P,x^*,X_1,X_2)$ \label{CC_build}
\\

$\phi \leftarrow \algoname{ComputeMaxFlow}(G,s,t)$  \label{CC_flow} \\
\If{ $F \ge \sum_{i\leq m-1}\Delta(P,x_i)+\sum_{x_i\in
X_2}\Delta(P,x_i)-(|Y| \cdot s(x^*,P_X)-T)$ \label{CC_if2}}
{return {\em true}}
return {\em false}

\end{algorithm}

\begin{prop}
Given a profile $P_X$ on $X$, a candidate $x^*\in X$ and a set of new candidates $Y$, a call to algorithm \algoname{CheckCowinner}$(P_X,x^*,Y)$
returns in polynomial time the answer \emph{true} \iff there exists an extension of $P_X$ in which $x^*$ is a cowinner.
\end{prop}

\beginproof
Algorithm \ref{algocc} starts by partitioning $X^\bullet$ into $X_1$ and $X_2$: an alternative $x \in X^\bullet$ is in $X_1$ if $|\text{HP}(P_X,x)|>\Delta(P_X,x_i)$ and in $X_2$ if $|\text{HP}(P_X,x)|\le \Delta(P_X,x)$. 

Let $x \in X_2$. Then 
by item (3) of Proposition~\ref{prop:opt}, for each vote in $V\in \text{HP}(P,x)$, we can safely put one candidate from $Y$ in the top position of $V$; this is done in the first phase of Algorithm \ref{algocc}, lines \ref{CC_for1} to \ref{CC_endfor1}.
Note that after adding a new candidate on top of a vote $V \in \text{HP}(P,x)$ and after updating $P$,
the modified vote will no longer belong to $\text{HP}(P,x)$. Instead, it will now belong to $\text{MP}(P,x')$
for some other candidate $x'$.

When Phase 1 is over, the score of $x \in X_2$ may still need to be lowered down, which can be done next by using votes from $\text{MP}(P_X,x)$. This is what Phase 2 does, from line \ref{CC_for2} to line \ref{CC_endfor2}.
There are three possibilities:

\begin{enumerate}
\item $|\text{HP}(P_X,x)| = \Delta(P_X,x)$. In this case, the votes in $\text{HP}(P_X,x)$ are sufficient to make $x^*$ catch up $x$: after Phase 1, we have $\Delta(P_X,x) = 0$ and Phase 2 is void; we are done with $x$.
\item $|\text{HP}(P_X,x)| < \Delta(P_X,x)$ and $|\text{HP}(P_X,x)| + |\text{MP}(P_X,x)| \geq \Delta(P_X,x)$: in this case, to make $x^*$ catch up $x$, it is enough to take $\Delta(P_X,x)-|\text{HP}(P_X,x)|$ arbitrary votes in $\text{MP}(P_X,x)$ and add one new candidate on top of them; this is what Phase 2 does, and after that we are done with $x$.
\item $|\text{HP}(P_X,x)| + |\text{MP}(P_X,x)| < \Delta(P_X,x)$: in this case, because of Proposition~\ref{prop:opt}, we know that it is safe to add one new candidate on top of {\em all} votes of $\text{MP}(P_X,x)$; this is what Phase 2 does; after that, we still need to lower down the score of $x$, which will require to add new candidates on top of votes of $\text{LP}(P_X,x)$.
\end{enumerate}

If at this point a newly added candidate has a score higher than $x^*$, then $x^*$ cannot win, and we can stop the program (line \ref{CC_existsy}). 

For readability, let us denote by $\widetilde{P}$ the profile obtained after Phases 1 and 2. For each
$x \in X_2$ satisfying condition 3, the only way to reduce $\Delta(\widetilde{P},x)$ is to put two
candidates of $Y$ within the top two positions in a vote 
of $\text{LP}(\widetilde{P},x)$, because in Phases 1 and 2 
we have used up all the votes in $\text{HP}(P,x)$ and $\text{MP}(P,x)$. 
Now, reducing $\Delta(\widetilde{P},x)$ by one unit will cost us two candidates in
$Y$, but meanwhile, $\Delta(\widetilde{P},x')$ is also reduced by one unit, where $x'$ is
the candidate ranked in the second position in $V$. We
must have $x'\in X_1$. We note that
$\bigcup_{x\in X_2}\text{LP}(P_X,x)\subseteq\bigcup_{x'\in X_1}\text{HP}(P_X,x')$.
Choosing optimally the votes in $\text{LP}(P_X,x)$ for each
$x\in X_2$ can be done by solving an integral max-flow instance which is build by algorithm \algoname{BuildMaxFlowGraph}
(note that in case where either $X_1$ or $X_2$ is empty, we just assume that the flow has a null value).

Let us show that $x^*$ is a possible cowinner if and only if the value of the flow from $s$ to $t$ is at least $\sum_{i\leq m-1}\Delta(\widetilde{P},x_i)+\sum_{x_i\in X_2}\Delta(\widetilde{P},x_i)-(|Y| \cdot s(x^*,P_X)-T)$. 
Observe that the flow does not necessarily bring all $\Delta(\widetilde{P},x_i)$ to 0, therefore we sometimes need a postprocessing 
consisting of adding further new candidates on top of some votes 
(see steps \nicobf{2 and 3} below).\\

Suppose first that the above max-flow instance has a solution whose value which is at
least $$\sum_{i\leq m-1}\Delta(\widetilde{P},x_i)+\sum_{x_i\in
X_2}\Delta(\widetilde{P},x_i)-(|Y| \cdot s(x^*,P_X)-T)$$
 We show how
to solve our cowinner problem from the solution to this flow problem. Because the instance is integral, there must exist an integral solution. We arbitrarily choose one integral solution $\phi$ (as returned by \algoname{ComputeMaxFlox}), which assigns to each
edge $(x_i,x_j)$ an integer $\phi(x_i,x_j)$ which represents the value of the flow on this edge. Here, we give a procedure
which produces an extension $P$ of $P_X$ where $x^*$ is a cowinner : 

\begin{enumerate}

\item For each $x_i\in X_2$ and each $V\in \text{LP}(\widetilde{P},x_i)$, if there
is a flow from $x_i$ to $x_j$ via $V$, then we obtain $V^*$ from $V$
by putting two candidates from $Y$ in the top positions (that is,
both $\Delta(\widetilde{P},x_i)$ and $\Delta(\widetilde{P},x_j)$ are reduced by $1$, which comes at the cost
of using candidates in $Y$ twice). It is possible since $|Y|\geq
2$.

\item For each $x_i\in X_2$, if
$\phi(x_i,t)<\Delta(\widetilde{P},x_i)$, then we arbitrarily choose $\Delta(\widetilde{P},x_i)-\phi(x_i,t)$ votes
$V\in \text{LP}(\widetilde{P},x_i)$ among those which haven't been
selected in the previous step, %such that $V^*$ is not defined in (1)
 and obtain $V^*$ by putting two candidates from $Y$ in the top two
positions (again, we will specify how to choose the two candidates
from $Y$ later). It is possible since $|Y|\geq 2$. 

\item For each $x_j\in X_1$, if $\phi(s,x_j)<\Delta(\widetilde{P},x_i)$, then we
arbitrarily choose $\Delta(\widetilde{P},x_i)-\phi(s,x_j)$ votes $V\in \text{HP}(\widetilde{P},x_j)$
such that $V^*$ is not defined above (in (1) or (2)), and then we
obtain $V^*$ by putting exactly one candidate from $Y$ in the top
position of $V$. This \jl{is possible because, by construction,} 
$|\text{HP}(\widetilde{P},x_j)|=|\text{HP}(P,x_j)|\geq \Delta(P,x_i)\geq  \Delta(\widetilde{P},x_i)$ for $x_j\in X_1$.

\item For each $V^*$, if a candidate $y \in Y$ is not selected for one of the first two 
positions, then it is ranked at the bottom position.
\end{enumerate}

In the above procedure (similarly to what is done in Algorithm \ref{algo:addnew}), priority is given to candidates from $Y$ with the lowest score (or the lowest index, in case of ties) when it comes to choose those to be added on top of the votes.

Let us now determine the number of times that new candidates from $Y$ are inserted on top of the votes.  Recall that 
until line \ref{CC_build} of the algorithm, we have used the
candidates from $Y$ exactly  $T$ times. 
Now consider the four-step procedure described above. 
Observe that to reduce by one unit the score deficit with respect to one candidate,  
 steps 1 and 3 require one occurrence of a candidate of $Y$ (step 1 uses two occurrences but reduces the score deficit with respect to two candidates), while step 2 requires \emph{two} occurrences.  
Thus, for each $i\leq m-1$, we have to
use $\Delta(\widetilde{P},x_i)$ times the candidates from $Y$, plus the additional occurrences required in step 2. 
More precisely, step 2 requires, for each $x_i \in X_2$, $\Delta(\widetilde{P},x_i) - \phi(x_i,t)$ additional occurrences of new candidates in the completed votes. Therefore, the total number of times that the
candidates of $Y$ are ranked either in first or second position (denoted $s_Y$ for readability), is such that: 
\begin{align}
s_Y \leq & \sum_{i\leq m-1}\Delta(\widetilde{P},x_i)+(\sum_{x_i\in
X_2}\Delta(\widetilde{P},x_i)-\sum_{x_i\in X_2} \phi(x_i,t))  \\
= & \sum_{i\leq
m-1}\Delta(\widetilde{P},x_i)+(\sum_{x_i\in
X_2}\Delta(\widetilde{P},x_i)-\phi) \label{Sy_1}
\end{align}

But we also have : 
\begin{align}
\phi \geq \sum_{i\leq
m-1}\Delta(\widetilde{P},x_i)+\sum_{x_i\in
X_2}\Delta(\widetilde{P},x_i)-(|Y| \cdot s(x^*,P_X)-T) \label{Sy_2}
\end{align}  
By combining (\ref{Sy_1}) and (\ref{Sy_2}), we thus get : 
\begin{align*}
s_Y \leq & |Y| \cdot s(x^*,P_X)-T \\
 \leq & |Y| \cdot s(x^*,P_X)
\end{align*}

That is, our algorithm will put candidates from $Y$ in the top two positions in
the extension no more than $|Y|\cdot s(x^*,P_X)$ times. Because the addition of new candidates is done in a cyclic order, each new candidate will eventually appear at most $s(x^*,P_X)$ in the top two positions of the votes. Thus, the score of these new candidates will not exceed that of $x^*$. It follows that $x^*$ is a cowinner \jl{in} $P^*$, since for all other candidates $x_i \in X$, we have $\Delta(P^*,x_i)\le 0$.

Next, we show that if $x^*$ is a possible \jl{cowinner}, then the value of a max-flow must be at least 
$$\sum_{i\leq
m-1}\Delta(\widetilde{P},x_i)+\sum_{x_i\in
X_2}\Delta(\widetilde{P},x_i)-(|Y| \cdot s(x^*,P_X)-T)$$ 
Due to
Proposition~\ref{prop:opt}, each extension profile $P^*$ of $P_X$
where $x^*$ becomes a cowinner to the problem instance can be
converted to a profile $\widetilde{P}$ as in the steps before line
  \ref{CC_build} in the algorithm. Now, for each $x_i\in X_2$, let
$l_i$ denote the number of votes $V\in \text{LP}(\widetilde{P},x_i)$ such
that in its extension $V^*$, the top two positions are the
candidates of $Y$. We must have that $l_i\geq
\Delta(\widetilde{P},x_i)$. For every $x_i\in X_2$, we arbitrarily
choose $l_i-\Delta(\widetilde{P},x_i)$ such votes, and move the
first ranked candidate to the bottom position. For each $x_j\in
X_1$, let $l_j$ denote the number of votes $V\in
\text{HP}(\widetilde{P},x_j)\cup \text{MP}(\widetilde{P},x_j)$ such that in its
extension $V^*$, a candidate from $Y$ is ranked in the top
position. We must have that $l_j\geq \Delta(\widetilde{P},x_j)$. For
every $x_j\in X_1$, we arbitrarily choose
$l_j-\Delta(\widetilde{P},x_j)$ such votes, and move the first
ranked candidate to the bottom position.

Now, let there be a flow from $x_j \in X_1$ to $x_i \in X_2$ via
$V$ if $V\in \text{LP}(\widetilde{P},x_i)$ and the top two positions
in $V^*$ are both in $Y$. This defines a flow whose value is at
least $\sum_{x_i\in X_2}\Delta(\widetilde{P},x_i)-\sum_{x_j\in
X_1}(l_j-\Delta(\widetilde{P},x_j))$. Because the score of each
candidate of $Y$ is no more than $s(x^*,P_X)$, we know that $|Y|
\cdot s(x^*,P_X)-T\geq \sum_{i\leq m-1}l_i$. Actually, $|Y|
\cdot s(x^*,P_X)$ is the maximum score that the whole set of new
candidates of $Y$ can reach in such a way that $x^*$ is a cowinner.
In the partial profile $\widetilde{P}$ (line \ref{CC_build} of Algorithm
\algoname{CheckCowinner}$(P_X,x^{*},Y)$), the global score of $Y$ is $T$.
Finally, since $\sum_{i\leq m-1}l_i+T$ corresponds to the global
score that $Y$ has in profile $P^*$ (where $x^*$ becomes a
cowinner), we get $|Y| \cdot s(x^*,P_X)\geq \sum_{i\leq m-1}l_i+T$. 

Hence, $|Y| \cdot s(x^*,P_X)-T\geq \sum_{i\leq m-1}l_i \geq
\sum_{x_i\in X_2}\Delta(\widetilde{P},x_i)+\sum_{x_j\in X_1}l_j$, or
equivalently, $-\sum_{x_j\in X_1}l_j\geq  \sum_{x_i\in
X_2}\Delta(\widetilde{P},x_i)-(|Y| \cdot s(x^*,P_X)-T)$. Hence, we get:
\begin{align*}
\phi \geq &\sum_{x_i\in X_2}\Delta(\widetilde{P},x_i)-\sum_{x_j\in X_1}(l_j-\Delta(\widetilde{P},x_j))\\
=&\sum_{i\leq m-1}\Delta(\widetilde{P},x_i)-\sum_{x_j\in X_1}l_j\\
\geq &\sum_{i\leq m-1}\Delta(\widetilde{P},x_i)+\sum_{x_i\in
X_2}\Delta(\widetilde{P},x_i)-(|Y|s(x^*,P_X)-T)
\end{align*}
Thus, we have shown that  $x^*$ is a possible cowinner if and only if the value of the flow from $s$ to $t$ is at least $\sum_{i\leq m-1}\Delta(\widetilde{P},x_i)+\sum_{x_i\in X_2}\Delta(\widetilde{P},x_i)-(|Y| \cdot s(x^*,P_X)-T)$. This concludes the proof. 
\endproof

\begin{coro}
\label{coro-2app}
Deciding whether $x^*$ is a possible cowinner for 2-approval with
respect to the addition of new candidates is in {\sf P}.
\end{coro}

To better understand Algorithm 1, we will now run it step by step on the example introduced previously.

\begin{example}
Consider the profile described in Example 1. We assume the number of new candidates is $k = 3$.
First, the initial scores of the candidates are $s(x^*,P_X) = 4$,  $s(x_1,P_X) = 7$,
$s(x_2,P_X) = 7$,  $s(x_3,P_X) = 8$,  $s(x_4,P_X) = 6$ and
$s(x_5,P_X) = 4$ and  $s(x_6,P_X) = 2$. The candidates 
whose score exceeds that of 
$x^*$ are $x_1$, $x_2$, $x_3$ and $x_4$, with the score
differences $\Delta(P,x_1) = 3$, $\Delta(P,x_2) = 3$, $\Delta(P,x_3)
= 4$ and $\Delta(P,x_4)  = 2$. At first phase, we check if there are
candidates $x_i$ for which $|HP(P_X,x_i)| \leq \Delta(P_X,x_i)$.
This is the case for $x_1$, $x_2$ and $x_3$, thus we put one new
candidate on top of $v_8$, $v_9$, $v_{13}$ and $v_{14}$. The updated
table is as follows:

$$\begin{array}{|l|l|l|l|l|}
\hline
& \mbox{\text{\rm HP}} & \mbox{\text{\rm MP}} & \mbox{\text{\rm LP}} & \Delta(P,x_i)\\ \hline
x_1 & & v_1 & v_5, v_6, v_7 & 1 \\
x_2 & & v_8', v_9' & v_{10}, v_{11}, v_{12} & 2 \\
x_3 & & v_{13}', v_{14}' & v_{15}, v_{16}, v_{17}, v_{18} & 3 \\
x_4 & v_5, v_6, v_{10}, v_{15}, v_{16} && v_{19} & 2\\ %\hline
\hline
\end{array}$$

Here, $v_i'$ refers to the vote $v_i$ to which new candidates have been added. 

At the second phase, $X_2 = \{x_1,x_2,x_3\}$ and $X_1 = \{x_4\}$ (we do not worry about $x_5$ and $x_6$ for which nothing special has to be done).
We put one new candidate on top of $v_1$, $v_8'$, $v_9'$, $v_{13}'$ and  $v_{14}'$, and we are done with $x_1$ and $x_2$
(since $\Delta(P,x_1)=0$ and $\Delta(P,x_2)=0$). The profile is now $\widetilde{P}$ and the updated table is : 

$$\begin{array}{|l|l|l|l|l|}
\hline
& \mbox{\text{\rm HP}} & \mbox{\text{\rm MP}} & \mbox{\text{\rm LP}} & \Delta(\widetilde{P},x_i)\\ \hline
x_3 & &  & v_{15}, v_{16}, v_{17}, v_{18} & 1\\
x_4 & v_5, v_6, v_{10}, v_{15}, v_{16} && v_{19} & 2 \\ %\hline
\hline
\end{array}$$

So far we have used the new candidates 9 times, and
$s(x^*,\widetilde{P}) = 4$, therefore if we have less than three new
candidates we stop ($x^*$ is not a possible cowinner) otherwise we
continue. Now the situation is as follows and we have to solve the
corresponding maxflow problem (we omit the value of edges when it
equals 1).

\begin{footnotesize}
\[
\begin{array}{ccccccccccccccccccc}
v_1 & v_2 & v_3 & v_4 & v_5 & v_6 & v_7 & v_8 & v_9 & v_{10} & v_{11} & v_{12} & v_{13} & v_{14} & v_{15} & v_{16} & v_{17} & v_{18} & v_{19} \\
\hline
\bullet & x_1 & x_2 & x_3 & x_1 & x_1 & x_1 & \bullet & \bullet & x_2 & x_2 & x_2 & \bullet & \bullet & x_3 & x_3 & x_3 & x_3 & x_4 \\
x^* & x^* & x^* & x^* & x_4 & x_4 & x_5 & \bullet & \bullet & x_4 & x_5 & x_5 & \bullet & \bullet & x_4 & x_4 & x_5 & x_6& x_6 \\
\vdots & \vdots &\vdots &\vdots &\vdots &\vdots &\vdots &\vdots &\vdots &\vdots &\vdots &\vdots &\vdots &\vdots &\vdots &\vdots &\vdots &\vdots &\vdots
\end{array}
\]
\end{footnotesize}

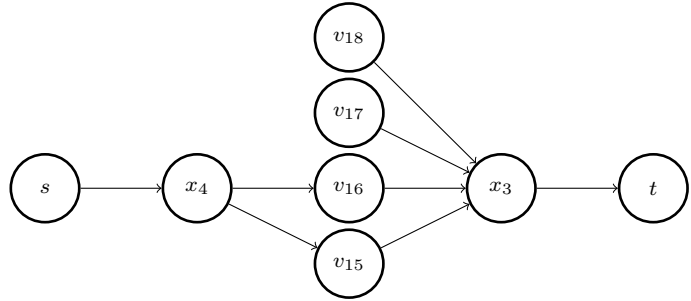
\begin{figure}[h!] \label{figureflow}
\begin{center}
\begin{tikzpicture}[scale=1]
\tikzstyle{s}=[circle,draw, line width=1pt, minimum size = 9mm, font= \footnotesize]

\node (s) at (0,2) [s] {$s$};
\node (x4) at (2,2) [s] {$x_4$};
\node (v15) at (4,1) [s] {$v_{15}$};
\node (v16) at (4,2) [s] {$v_{16}$};
\node (v17) at (4,3) [s] {$v_{17}$};
\node (v18) at (4,4) [s] {$v_{18}$};
\node (x3) at (6,2) [s] {$x_3$};
\node (t) at (8,2) [s] {$t$};

\draw [->] (s) -- (x4);
\draw [->] (x4) -- (v15);
\draw [->] (x4) -- (v16);
\draw [->] (v15) -- (x3);
\draw [->] (v16) -- (x3);
\draw [->] (v17) -- (x3);
\draw [->] (v18) -- (x3);
\draw [->] (x3) -- (t);

\end{tikzpicture}
\caption{The flow graph returned by \algoname{BuildMaxFlowGraph}$(\widetilde{P},x^*,\{x_4\},\{x_3\})$.} \label{flow}
\end{center}
\end{figure}

The maximum flow has value 1 and is obtained for instance by having a
flow 1 for instance through the edges $s \rightarrow x_4$, $x_4
\rightarrow v_{16}$, $v_{16} \rightarrow x_{3}$,  $x_{3} \rightarrow
t$ (going through $v_{15}$ is an equally good option). Therefore we
place two new candidates on top of $v_{16}$, which has the effect of making
the score of $x_3$ and $x_4$ decrease by one unit each. We still
have to make the score of $x_4$ decrease by one unit, and for this
we must place one new candidates on top of any of the votes $v_5$, $v_6$,
$v_{10}$, $v_{15}$ (say $v_5$). In total we will have used the new
candidates 12 times, therefore, $c$ is a possible cowinner if and
only if the number of new candidates is at least 3. A possible
extension (with 3 new candidates) is as follows:

\begin{footnotesize}
\[
\begin{array}{ccccccccccccccccccc}
v_1 & v_2 & v_3 & v_4 & v_5 & v_6 & v_7 & v_8 & v_9 & v_{10} & v_{11} & v_{12} & v_{13} & v_{14} & v_{15} & v_{16} & v_{17} & v_{18} & v_{19} \\
\hline
{\bf y_1} & x_1 & x_2 & x_3 & {\bf y_2} & x_1 & x_1 & {\bf y_1} & {\bf y_2} & x_2 & x_2 & x_2 & {\bf y_1} & {\bf y_2} & x_3 & {\bf y_1} & x_3 & x_3 & x_4 \\
x^* & x^* & x^* & x^* & x_1 & x_4 & x_5 & {\bf y_3} & {\bf y_3} & x_4 & x_5 & x_5 & {\bf y_3} & {\bf y_3} & x_4 & {\bf y_2} & x_5 & x_6 & x_6 \\
\vdots & \vdots &\vdots &\vdots &\vdots &\vdots &\vdots &\vdots &\vdots &\vdots &\vdots &\vdots &\vdots &\vdots &\vdots &\vdots &\vdots &\vdots &\vdots
\end{array}
\]
\end{footnotesize}

\end{example}

\subsection{$K$-approval, two new candidates}\label{kapp-2}

Let $X = \{x^*\} \cup \{x_1,\ldots, x_p\}$ be the set of (initial) candidates, $x^*$ being the candidate that we want to make a cowinner,
$Y=\{y_1,y_2\}$ the two new candidates, and $P_X = \langle V_1,\ldots, V_n\rangle$ the initial profile,
where each $V_i$ is a sequence of $K$ candidates in $X$.
We first introduce the following notation:

\begin{itemize}

\item For each $x \in X$, $U^{P_X}(x)$ is the number of votes in $P_X$ whose candidates ranked $K-1$ and $K$ are respectively $x$ and $x^*$, and
$T^{P_X}(x) = S_{K-2}(x,P_X) + U^{P_X}(x)$. (Recall that
$S_{K-2}(x,P_X)$ is the number of voters in $P_X$ who rank $x$ in
the first $K-2$ positions.)
\end{itemize}

We establish the following lemma.

\begin{lemma}\label{lemma-2app}
For each $x \in X$, there exists a completion $Q$ of $P_X$ by adding
two candidates such that $S_K(x,Q) \leq S_K(x^*,Q)$ if and only if
$T^{P_X}(x) \leq S_K(x^*,P_X)$.
\end{lemma}

\beginproof
Assume $T^{P_X}(x) > S_K(x^*,P_X)$, and let $Q$ be  a completion of
$P_X$ by adding two candidates \jl{in which $x^*$ is a cowinner}. Let us partition $P_X$ into $P_1$,
$P_2$ and $P_3$, as follows: every vote in $P_1$ is such that the
candidates ranked $K-1$ and $K$ are respectively $x$ and $x^*$;
$P_2$ contains all votes ranking $x$ in the first $K-2$ positions;
and $P_3$ contains all other votes in $P_X$. Let $Q_1$, $Q_2$ and
$Q_3$ be the corresponding votes in $Q$, and let $\alpha$ be the
number of votes in $Q_1$ where the  two new candidates have been placed in 
the first $K$ positions, thus eliminating both $x$
and $x^*$ from the $K$ first positions; clearly, we have $S_K(x,Q_1)
= S_K(x,P_1)-\alpha = U^{P_X}(x) - \alpha$ and $S_K(x^*,Q_1) \leq
S_K(x^*,P_1)-\alpha$ (the inequality can be strict, in case there
are some votes in $Q_1$ where only one new candidate was placed
in the first $K$ positions). Now, regardless of the position of
the two new candidates, we have $S_K(x,Q_2) = S_{K-2}(x,P_2)$. We
get $S_K(x,Q) = S_K(x,Q_1) + S_K(x,Q_2)  + S_K(x,Q_3) \geq
U^{P_X}(x) - \alpha + S_{K-2}(x,P_2) = T^{P_X}(x) - \alpha$, whereas
$S_K(x^*,Q) \leq S_K(x^*,P_X) - \alpha$. The initial assumption
$T^{P_X}(x) > S_K(x^*,P_X)$ implies $T^{P_X}(x)- \alpha > S_K(x^*,P_X)
- \alpha$, therefore $S_K(x,Q) > S_K(x^*,Q)$.

Conversely, assume $T^{P_X}(x) \leq  S_K(x^*,P_X)$, and let us build
$Q$ as follows: we introduce one new candidate on top of each vote
of $P_X$ that ranks $x$ in position $K$, and two new candidates on
top of each vote of $P_X$ that ranks $x$ in position $K-1$ and $x'
\neq x^*$ in position $K$. It is easy to check that $S_K(x^*,Q) =
S_K(x^*,P_X)$. Now, the only votes of $Q$ where $x$ remains among
the first $K$ position are those of $Q_1$ and of $Q_2$, therefore
$S_K(x,Q) = T^{P_X}(x)  \leq  S_K(x^*,P_X) = S_K(x^*,Q)$.
\endproof

\begin{prop}\label{prop-2cand}
Deciding whether $x^*$ is a possible cowinner for $K$-approval with
respect to the addition of 2 new candidates is in {\sf P}.
\end{prop}

\beginproof
A consequence of Lemma \ref{lemma-2app} is that if $T^{P_X}(x) >
S_K(x^*,P_X)$ for some $x$, then $x^*$ cannot be a possible cowinner in $P_X$
under $2$-approval with 2 new candidates; and obviously, checking
whether  $T^{P_X}(x) > S_K(x^*,P_X)$ holds for some $x$ can be done
in polynomial time. Therefore, {\em from now on, we assume that
$T^{P_X}(x) \leq S_K(x^*,P_X)$ holds for every $x \in X$} ---
assuming this will not change the complexity of the problem.

We now give a polynomial reduction from the possible cowinner
problem for $K$-approval and $2$ new candidates to the possible
cowinner problem for $2$-approval and $2$ new candidates, which we
already know to be polynomial. Let
$\langle N, X, P_X, 2\rangle$ 
be an instance of the possible cowinner problem for $K$-approval
with respect to the addition of $2$ new candidates.
We build an instance $\langle N', X', R_{X'}, 2\rangle$ 
of the possible cowinner problem for $2$-approval
with respect to the addition of $2$ new candidates in the
following way. The profile $P_X$ is translated into the following
profile $R=R_{X'}$:

\begin{itemize}
\item the set of candidates is $X'=X\cup \{z_j, 1 \leq j
\leq  \sum_{x \in X \setminus \{x^*\}}S_{K-2}(x,P_X)\} \cup \{z_j', 1
\leq j \leq S_{K-2}(x^*,P_X)\}$, where all $z_j$ and $z_j'$ are
fresh candidates;

\item for every vote $V_i$ in $P_X$, we have in $R$ a vote $W_i$ including the candidates
ranked in positions $K-1$ and $K$ of $V_i$, and then the remaining candidates in any order. We denote by $R_1$ be the resulting set of votes;

\item for every $x \in X\setminus \{x^*\}$, we have $S_{K-2}(x,P_X)$ votes $x z_j$, and then the remaining candidates in any order.
We denote by $R_2$ the resulting set of votes;

\item similarly, we have $S_{K-2}(x^*,P_X)$ votes $z_j' x^*$, and then the remaining candidates in any order.
We denote by $R_3$ the resulting set of votes.
\end{itemize}

We note that if $x \in X$ then $S_{K}(x,P_X)=S_2(x,R)$, and for
every fresh candidate $z$, $S_2(z,R)=1$. Without loss of generality
we assume $S_K(x^*,P_X) \geq 1$ (otherwise we know for sure that
$x^*$ cannot be a possible cowinner).

We decompose the rest of the proof into two lemmas.

\begin{lemma}\label{lemmaprop5-1} If $x^*$  is a possible cowinner for $K$-approval with 2 new candidates  in $P_X$ , then it is is a possible cowinner for 2-approval with 2 new candidates in $R$.
\end{lemma}

\beginproof
Suppose that $x^*$ is a possible cowinner for $K$-approval with 2
new candidates $Y=\{y_1,y_2\}$ in $P_X$ and let $P' = \langle V_1',
\ldots, V_n'\rangle$ be an extension of $P_X$ with two new
candidates where $x^*$ is a cowinner. Let us use these two new
candidates in the same way in $R$: every time a new candidate is
used for being placed on top of $V_i$, it is also used for being placed on top of $W_i$. Let
$R'$ be the resulting profile. All candidates in $X$ have the same
scores in $P_X$ and in $R$, they also will have the same scores in
$P'$ and $R'$; as for the fresh candidates $z_j,z'_j$, $S_2(z_j,R')
=S_2(z'_j,R')= 1 \leq S_2(x^*,R')$; therefore, $x^*$ is a cowinner
in $R'$ and a possible cowinner for 2-approval with 2 new candidates
in $R$.
\endproof

\begin{lemma}\label{lemmaprop5-2} If $x^*$  is a possible cowinner for $K$-approval with 2 new candidates in $R$ , then it is a possible cowinner for 2-approval in $P_X$.
\end{lemma}

\beginproof
Suppose that $x^*$ is a possible cowinner  for
$2$-approval with 2 new candidates $Y=\{y_1,y_2\}$ in $R$, and let
$R'$ be a completion of $R$ where $x^*$ is a cowinner for
$2$-approval. Let us write $R' = R_1' \cup R_2' \cup R_3'$, where
$R_1'$ (resp. $R_2'$, $R_3'$) consists in the completions of the
votes in $R_1$ (resp. $R_2$, $R_3$). By a slight abuse of language
we denote by $R_1, R_1'$ etc. only the part of the votes in
$R_1,R_1'$ etc. consisting of the top two candidates only.

We first claim that we can assume without loss of generality that
$R_2' = R_2$ and $R_3' = R_3$ that is, the only votes in $R'$ where
some new candidates have been placed on one of the top two positions 
are in $R_1'$. Suppose this is not the case; then we are in one of
the following four situations: (1) there is a vote in $R_2'$ of the
form $y_j x_i$, where $y_j \in Y$ and $x_i \in X$, or (2)  there is
a vote in $R_2'$ of the form $y_1 y_2$ or $y_2 y_1$, or (3)  there
is a vote in $R_3'$ of the form $y_i z'_j$ or (4)  there is a vote
in $R_3'$ of the form $y_1 y_2$ or $y_2 y_1$.
Consider first cases (1), (3) and (4). Take one of these votes in $R_2'$ (case (1)) or in $R_3'$ (cases (3) or (4))
and replace it by the original vote $x z_j$ in $R_2$ (case 1) or in $z'_j x^*$ in $R_3'$ (cases (3) or (4)). Let $R''$ be the profile obtained. We have %\\
%\begin{tabular}{l}
$S_2(x^*,R'') \geq S_2(x^*,R') \geq 1$,  %\\
for every $x_i \in X$, $S_2(x_i,R'') = S_2(x_i,R')$,  %\\
for every $z_j$, $S_2(z_j,R'') \leq 1$,  %\\
and for every $z_j'$, $S_2(z'_j,R'') \leq 1$.

%\end{tabular}\\
Therefore, when transforming $R'$ into $R''$,  the score of $x^*$
does not decrease whereas the score of all other candidates does not
increase; because $x^*$ is a cowinner in $R'$, it is still a
cowinner in $R''$. Lastly, $R''$ is also an extension of $R$.

By induction, if we perform this operation for each occurrence of
cases (1), (3) or (4), we end up with a profile $R''$, which is an
extension of $R$ for which situations (1), (3)  and (4) do not
occur, and such that $x^*$ is a cowinner for 2-approval in $R''$. 
%the 2-approval winner for $R''$ is $x^*$. 
Let $R'' = R_1'' \cup R_2'' \cup R''_3 = R_1'' \cup R_2'' \cup R_3$.

Now, consider case (2). Let $x_i z_j$ be one of the votes in $R_2$
corresponding to a vote $y_1 y_2$ (or $y_2 y_1$) in $R_2''$. Apply
the following procedure in this order:

\begin{enumerate}

\item Assume that $x_i$ does not appear in $R_1''$ except in votes of the form $x_i x^*$, and let $R'''$
be the profile obtained from $R''$ by replacing the vote $y_1 y_2$
in $R_2''$ by the original vote  $x_i z_j$ in $R_2$. Then
$S_2(x_i,R''') = S_2(x_i,R_1''') + S_2(x_i,R_2''') \leq
S_2(x_i,R_1''') + S_2(x_i,R_2)$. Now, $S_2(x_i,R''') \leq S_2(x_i,R''') = U^{P_X}(x_i)$ and $S_2(x_i,R_2) = S_{K-2}(x_i,P_X)$, therefore $S_2(x_i,R''') \leq$  $U^{P_X}(x_i) + S_{K-2}(x_i,P_X) =
T^{P_X}(x_i) \leq S_K(x^*,P_X) = S_2(x^*,R) = S_2(x^*,R''')$.
Therefore, $x^*$ is also a cowinner in $R'''$.

\item Now, assume that $x_i$ appears in at least one vote of $R_1''$ of the form $x^* x_i$, $x_i x_j$ or $x_j x_i$.
If this is a vote of the form $x_i x_j$, we replace $y_1 y_2$ in
$R_2''$ by the original vote $x_i z_j$ in $R_2$ and the vote $x_i
x_j$ by a vote $y_1 y_2$. If this is a vote of the form $x^* x_i$ or
$x_j x_i$, we replace $y_1 y_2$ in $R_2''$ by the original vote $x_i
z_j$ in $R_2$ and the vote $x^* x_i$ (resp. $x_j x_i$) by $y_1 x^*$
(resp. $y_1 x_j$). In all three cases, the score of all candidates
remain the same after the transformation, except the score of $y_2$
and $z_j$, which can only decrease, therefore $x^*$ is still a
cowinner after the transformation.

\end{enumerate}

We perform this procedure on $x_i$ iteratively until all the votes  $y_1 y_2$ (or $y_2 y_1$) in $R_2'''$
have been replaced by the original votes  $x_i z_j$ in $R_2$. After doing this sequentially on all candidates
of $X$ such that case (2) occurs, we end up with a profile $R''''$ of $P$ such  that $R_2'''' = R_2$ and $R_3'''' = R_3$ and $x^*$ is a cowinner in $R''''$.
This proves the claim.\medskip

Now, let $R'$ be a completion of $R$ where $x^*$ is a cowinner for $2$-approval, where $R_2' = R_2$ and $R_3' = R_3$.
From $R'$ we build the following extension $P'$ of $P$: for every vote $W_i \in R_1$,
\begin{itemize}
\item if $W_i'$ is of the form $x^* x'$ then $V_i' = W_i'$;
\item if $W_i'$ is of the form $y_i x$ then $V_i'$ is obtained from $V_i$ by placing $y_i$ on top; 
\item if $W_i'$ is of the form $y_1 y_2$ (or $y_2 y_1$) then $V_i'$ is obtained from $V_i$ by placing $\{y_1, y_2\}$ on top. 
\end{itemize}
The scores of all candidates are the same in $P'$ and in $R'$, therefore $x^*$  is a cowinner in $P'$ if only if it is a cowinner in $R'$.
Therefore, it is a cowinner in $P'$, which means that  $x^*$ is a possible cowinner for 2-approval in $P$.
\endproof

We can now end the proof of Proposition \ref{prop-2cand}: from Lemmas \ref{lemmaprop5-1} and \ref{lemmaprop5-2}  we conclude that deciding whether $x^*$ is a possible cowinner for $K$-approval with respect to the addition of two candidates can be polynomially reduced to a problem of a deciding whether $x^*$ is a possible cowinner for $2$-approval, which we know is in {\sf P}.
\endproof

\subsection{$3$-approval, $3$ new candidates}\label{kapp-2}
We will now see that the problems addressed in previous subsections constitute the frontier of what can be solved in polynomial-time for $K$-approval rules.
In the rest of this paper, the hardness proofs will use reductions from the 3-dimensional matching (\textsc{3-DM}) problem.
\begin{defin}\label{def:3DM}
An instance of \textsc{3-DM} consists of a subset
$\mathcal{C}=\{e_1,\dots,e_m\}\subseteq A\times B\times C$ of
triples, where $A,B,C$ are 3 pairwise disjoint sets of size $n'$ with
$A=\{a_1,\dots,a_{n'}\}$, $B=\{b_1,\dots,b_{n'}\}$ and
$C=\{c_1,\dots,c_{n'}\}$. 
For $z\in A\cup B\cup C$, $d(z)$
denotes the number of occurrences of $z$ in $\mathcal{C}$, that is
the number of triples of $\mathcal{C}$ which contain $z$. 
A matching is a subset
$M\subseteq\mathcal{C}$ such that no two elements in $M$ agree on
any coordinate.
The \textsc{3-DM} problem consists in answering this question: does there
exist a perfect matching $M$ on $\mathcal{C}$, that is, a matching
of size $n'$? 
\end{defin}

The  \textsc{3-DM}  problem is known to be
\np-complete (problem [SP1] page 221 in \cite{GJ79}), 
even with the restriction where $\forall z\in A\cup B\cup C$, $d(z)\in\{2,3\}$ (that is, no element of
$A\cup B\cup C$ occurs in more than 3 triples, and each element of
$A\cup B\cup C$ appears in at least 2 triples). 

\begin{prop}\label{PropNP-completeApproval}
Deciding if $x^*$ is a possible cowinner for 3-approval with respect
to the addition of 3 new candidates, is an \np-complete problem.
\end{prop}

\beginproof
This problem is clearly in \np. The hardness proof is based on a reduction
from \textsc{3-DM} (see Definition \ref{def:3DM}).

Let  $I=(\mathcal{C},A\times B\times C)$ be an instance of
\textsc{3-DM} with $n'\geq 3$ and $\forall z\in A\cup B\cup C$, $d(z)\in\{2,3\}$. 
From
$I$, we build an instance of the PcWNC problem as follows. The set
$X$ of candidates contains $x^*$, $X_1= \{x'_i,y'_i,z'_i:1\leq i\leq
n'\}$ where $x'_i,y'_i,z'_i$ correspond to elements of $A\cup B\cup
C$ and a set $X_2$ of dummy candidates. We now describe the votes
informally; their formal definition will follow.
The set $N$ of voters
contains $N_1=\{v^e:e\in\mathcal{C}\}$ and a set $N_2$ of dummy
voters. For each voter, we only indicate her \jl{first three} candidates.
Thus, the vote of $v^e$ is $(x'_i,y'_j,z'_k)$ where
$e=(a_i,b_j,c_k)\in\mathcal{C}$. The preference of dummy voters are
such that : 
\begin{itemize}
\item [$(i)$] the \jl{scores of the candidates in $X$ satisfy} $\forall
x\in X_1$, $S_{3}(x,P_X)=n'+1$, $S_{3}(x^*,P_X)=n'$ and $\forall
x\in X_2$, $S_{3}(x,P_X)=1$; 
\item[$(ii)$] the vote of any voter of $N_2$ contains
at most one candidate from $\{x'_i,y'_i,z'_i:1\leq i\leq n'\}$ in
the first three positions, and if it contains one, then it is in top
position.
\end{itemize}

Formally, the instance $\langle N, X, P_X, 3, x^*\rangle$ 
of the possible cowinner problem for $3$-approval and $3$ new candidates
is described as follows: the set
of voters is $N=N_1\cup N_2$ where $N_1=\{v^e:e\in\mathcal{C}\}$ and
$N_2=N_A\cup N_B\cup N_C\cup N_{x^*}$, the set of candidates is
$X=X_1\cup X_2$. For the candidates in
$X$, we have $x^*$ together with : 

\begin{itemize}

\item $X_1=X'\cup
Y'\cup Z'$ where $X'=\{x'_1,\dots,x'_{n'}\}$, $Y'=\{y'_1,\dots,y'_{n'}\}$
and $Z'=\{z'_1,\dots,z'_{n'}\}$.

\item $X_{2}= %\{x^*\}\cup
 \{x^*_i: 1\leq i\leq 2n'\}\cup \{x_i^j:1\leq i\leq n'$, $1\leq j\leq 2\left(n'-d(x_i)+1\right)\}\cup
\{y_i^j:1\leq i\leq n'$, $1\leq j\leq 2\left(n'-d(y_i)+1\right)\}\cup
\{z_i^j:1\leq i\leq n'$, $1\leq j\leq 2\left(n'-d(z_i)+1\right)\}$.

Note that $n'-d(x_i)+1\geq 1$ since $d(z)\leq 3\leq n'$.
\end{itemize}

For each voter $v_i\in N$, we only indicate her \jl{first three} candidates
(in the order of preference). The set of all $X$-votes $\P_X$ of the voters in $N$ is
as follows : 
\begin{itemize}
\item $N_A=\{v_{i,j}^A:1\leq i\leq n'$,
$0\leq j \leq \left(n'-d(x_i)\right)\}$. The vote of $v_{i,j}^A$ is
$(x'_i,x_i^{2j+1},x_i^{2j+2})$.

\item $N_B=\{v_{i,j}^B:1\leq i\leq n'$,
$0\leq j \leq \left(n'-d(y_i)\right)\}$.  The vote of $v_{i,j}^B$ is
$(y'_i,y_i^{2j+1},y_i^{2j+2})$.

\item  $N_C=\{v_{i,j}^C:1\leq i\leq n'$,
$0\leq j \leq \left(n'-d(z_i)\right)\}$. The vote of $v_{i,j}^C$ is
$(z'_i,z_i^{2j+1},z_i^{2j+2})$.

\item  $N_1=\{v^e:e\in\mathcal{C}\}$.  The vote of
$v^e$ is $(x'_i,y'_j,z'_k)$ where $e=(a_i,b_j,c_k)\in\mathcal{C}$.

\item $N_{x^*}=\{v_j^{x^*}:0\leq j \leq n'-1\}$.  The vote of
$v_j^{x^*}$ is $(x^*,x^*_{2j+1},x^*_{2j+2})$.
\end{itemize}

We claim that $I$ admits a perfect matching $M\subseteq \mathcal{C}$
if and only if $x^*$ becomes a possible cowinner by adding three new
candidates.

Let $Y=\{y_1,y_2,y_3\}$ be the new candidates added. Since we cannot
increase the score of $x^*$, we must decrease by one point the scores
of candidates of $X'\cup Y' \cup Z'$. Let us focus on candidates in
$X'$. In order to reduce the score of $x'_i$, we must modify the
votes of voters in $N_1$ or in $N_A$. By construction, each such
voter must put $y_1,y_2,y_3$ in the first three positions (since in $N_A$ or
from $(ii)$, candidates of $X'$ are put in top position when they
appear in the first three positions) and then, the score of each $y_i$ increases
by 1 at each time. Since there are $n'$ candidates in $X'$, we deduce
$S_{3}(y_i,P)\geq n'$ for every $i=1,2,3$. On the other hand, if
$x^*$ becomes a cowinner, $S_{3}(y_i,P)\leq S_{3}(x^*,P)\leq
S_{3}(x^*,P_X)=n'$ from $(i)$. Thus, $S_{3}(y_i,P)= n'$ for every
$i=1,2,3$ and there are exactly $n'$ voters $N'$ which put
$y_1,y_2,y_3$ in the first three positions (for the remaining voters of
$N\setminus N'$, $y_i$ is ranked
in position at least 4 for every $i=1,2,3)$.\\

We claim that $N'\subseteq N_1$. Otherwise, at least one voter of
$N_A$ put $y_1,y_2,y_3$ in the first three positions. There remains at
most $n'-1$ voters of $N'$ to decrease  by 1 the score of candidates
in $Y'$. It is impossible because $|Y'|=n'$ and, from $(ii)$ and by
construction of $N_1$, each candidate of $Y'$ appears at most once
in the first three positions for all voters. Finally, since the score of
candidates in $Y'\cup Z'$ must also decrease by 1, we deduce that
$x^*$ is a possible cowinner iff $M=\{e\in \mathcal{C}:$
$y_1,y_2,y_3$ are in the first three positions for voter $v_e\}$ is a
perfect matching of $\mathcal{C}$.
\endproof

\subsection{General case}

We finalize the study of the possible cowinner problem for $K$-approval with respect to  candidate addition  by showing that the problem is hard in any other case. For this we proceed in two steps: we first prove that for each $k\geq 3$,
the problem {\sc PcWNC}($k$) 
for $3$-approval 
is \npc\ (Lemma \ref{Kappk>3}). Next we prove that if the problem {\sc PcWNC}($k$) 
for $K$-approval 
is \npc\, then it is also the case for the problem {\sc PcWNC}($k$) 
for $(K+1)$-approval (Lemma \ref{KappK+1}).

\begin{lemma}\label{Kappk>3}
For all $k \geq 3$, the problem {\sc PcWNC}$(k)$ for $3$-approval can be reduced in polynomial-time to the 
problem {\sc PcWNC}$(k+1)$ for $3$-approval.
\end{lemma}

\beginproof
Let $\langle N, X,P_X, k, x^*\rangle$, where $P =P_X= \langle V_1, \ldots, V_n\rangle$, 
be an instance of $\text{\sc PcWNC}(k)$ 
for $3$-approval. 
Assume $S_3(x^*,P)\geq 1$ (otherwise, the problem is trivial). Consider the following instance
$\langle N',X',Q_{X'}, k+1, x^*\rangle$ 
of the $\text{\sc PcWNC}(k+1)$ for $3$-approval: 

\begin{itemize}
\item the set of candidates is $X' = X \cup \{z\} \cup \{t_i^1,t_i^2 \ | \ 1 \leq i \leq  2S_3(x^*,P)\}$; 
\item there are $n+2S_3(x^*,P)$ votes:
\begin{itemize}
\item for every vote $V_j$ in $P$ we have a vote $W_j$ in $Q$ whose first three candidates are
 the same as in $V_j$ and in the same order, and the other candidates are in an arbitrary order.
\item for every $i = 1, \ldots, 2  S_3(x^*,P)$, we have a vote $U_i$ in which the first 3 candidates
 are $t_i^1, t_i^{2}, z$, the remaining candidates being ranked arbitrarily.
\end{itemize}
\end{itemize}

Assume $x^*$ is a possible cowinner for $P=P_{X}$ (w.r.t. the
addition of $k$ new candidates) and let $P'$ be an extension of $P$
where $x^*$ is a cowinner. Let  $Y=\{y_1,\ldots, y_k\}$ denote the
new candidates for the instance $\langle N,X, P_X, k\rangle$, and
$Y'=\{y_1,\ldots, y_{k+1}\}$ the new candidates for the instance
$\langle N', X', Q_{X'}, k+1\rangle$. Consider the following
extension $Q'$ of $Q=Q_{X'}$: for every vote $V'_j$ of $P'$ we have
a vote $W'_j$ in $Q'$ whose 3 first candidates are the same as in
$V'_j$ (and the remaining ones in an arbitrary order); and for every
vote $U_i$ such that $1 \leq i \leq S_3(x^*,P)$ we have a vote
$U'_i$ whose first 3 candidates are $y_{k+1}, t_i^1, t_i^{2}$ and
for every vote $U_i$ such that $S_3(x^*,P)+1 \leq i \leq 2 S_3(x^*,P)$, we have a vote
$U'_i$ whose first 3 candidates are $t_i^1, t_i^{2}, z$. It is easy
to check that $Q'$ is an extension of $Q$. The scores of all
candidates in $X \cup Y$ are the same in $P'$ and $Q'$, while the
score of each $t_i^1,t_i^2$ is 1, the scores of $z$ and of $y_{k+1}$
are $S_3(x^*,P)$; therefore $x^*$ is a cowinner in $Q'$ and a
possible cowinner in $Q$.

Conversely, assume $x^*$ is a possible cowinner in $Q=Q_{X'}$ and
let $Q'$ be an extension of $Q$ in which $x^*$ is a cowinner.
We are now going to reason abut the number of occurrences
of the new candidates $y_1,\ldots, y_{k+1}$ in the first three positions of the votes of $Q'$.
For the sake of notation, for any vote $V$ we denote $S_3(Y',V) = \sum_{y \in Y'}S_3(y',V)$:
in words,  $S_3(Y',V)$ is the number of new candidates in the first three positions of $V$.
Similarly,  if $R$ is a profile, we denote $S_3(Y',R)= \sum_{y \in Y'}S_3(y',R)$.
 
Without loss of generality, we assume that $S_3(x^*,Q') = S_3(x^*,Q)
= S_3(x^*,P)$, since under $3$-approval  it is never beneficial to
decrease the score of $x^*$ to make it a possible cowinner.
We have $S_3(z,Q') \leq S_3(x^*,Q') = S_3(x^*,P)$ and $S_3(z,Q) = 2 
S_3(x^*,P)$, therefore a new candidate must be put above $z$ in at
least $S_3(x^*,P)$ votes $U'_i$; therefore, 
$$\sum_{j=1}^{2S_3(x^*,P)}S_3(Y',U_i) \geq S_3(x^*,P) ~~~(1)$$
Now, $S_3(Y',Q') = \sum_{i=1}^n S_3(Y',W_i) +\sum_{j=1}^{2S_3(x^*,P)}S_3(Y',U_i)$, which together with (1) entails
$$\sum_{i=1}^n S_3(Y',W_i) \leq S_3(Y',Q') - S_3(x^*,P) ~~~(2)$$
Now, $x^*$ is a cowinner in $Q'$, therefore, for all $y_j \in Y'$ we have $S_3(y_j,Q') \leq S_3(x^*,Q') = S_3(x^*,P)$, from which we get
$$S_3(Y',Q') \leq (k+1)S_3(x^*,P) ~~~(3)$$
From (2) and (3) we get
$$\sum_{i=1}^n S_3(Y',W_i) \leq kS_3(x^*,P) ~~~(4)$$
Now, consider the extension $P'$ of $P$ built from the restriction of $Q'$ to $\{W_1', \ldots, W_n'\}$ by changing the candidates in $Y$ placed in the first three positions {\em in such a way that each candidate appears at most in $S_3(x^*,P)$ votes}, which is made possible by (4). We have:

\begin{itemize}
\item $S_3(x^*,P') = S_3(x^*,P')$; 
\item for each $y \in Y$, $S_3(y,P') \leq S_3(x^*,P) = S_3(x^*,P')$;
\item for each $x \in X \setminus \{x^*\}$, $S_3(x,P') = S_3(x,Q')$; because $x^*$ is a possible cowinner in $Q'$, we have $S_3(x,Q') \leq S_3(x^*,Q') = S_3(x^*,P)$, therefore, 
$S_3(x,P') \leq S_3(x^*,P) =   S_3(x^*,P')$.
\end{itemize} 
From this we conclude that $x^*$ is a possible cowinner in $P'$. 
\endproof

\begin{lemma}\label{KappK+1}
The problem $\text{\sc PcWNC}(k+1)$ for $K$-approval 
can be reduced in polynomial-time to the problem $\text{\sc PcWNC}(k)$ for $(K+1)$-approval.
\end{lemma}

\beginproof
Let $\langle N, X,P_X,
k, x^*\rangle$ where $P_X = \langle V_1, \ldots, V_n\rangle$ 
be an instance of PcWNC($k$) 
for $K$-approval. 
Consider the following instance $\langle N',X',R_{X'},k, x^*\rangle$ 
of the PcWNC($k$) for $(K+1)$-approval: 

\begin{itemize}
\item the set of candidates is $X' = X \cup \{t_i \ | \ 1 \leq i \leq n\}$;
\item for every vote $V_i$ in $P$ we have a vote $W_i$ in $R$ whose top candidate is $t_i$
and the candidates ranked in position 2 to $K+1$ are the candidates ranked in positions 1 to $K$ in $V_i$,
the remaining candidates being ranked arbitrarily.
\end{itemize}

Assume $x^*$ is a possible cowinner for $P=P_X$ and let $P' =
\langle V_1', \ldots, V_n'\rangle$ be an extension of $P$  where
$x^*$ is a cowinner. Denote by $y_1, \ldots, y_k$ the new
candidates. Consider the extension $R' = \langle W_1', \ldots,
W_n'\rangle$ of $R=R_{X'}$ where $W_i'$ ranks $t_i$ first and then
the candidates ranked in the first $K$ positions in $V_i'$. For
every $x \in X$ we have $S_{K+1}(x,R') = S_{K}(x,P')$; for every $i
= 1,\ldots, k$ we have $S_{K+1}(y_i,R') = S_{K}(y_i,P')$; and for
every $j = 1, \ldots, n$, we have $S_{K+1}(t_j,R') = 1$. Therefore
$x^*$ is a possible cowinner in $R'$ and a possible cowinner in
$R$.

Conversely, assume $x^*$ is a possible cowinner in $R=R_{X'}$ and
let $R' = \langle W_1', \ldots, W_n'\rangle$ be a completion of $R$
in which it is a possible cowinner. Since none of the $t_i$
threatens $x^*$, without loss of generality we assume $t_i$ still
appears in the first $K+1$ positions of $W_i'$---otherwise,
change $W_i'$ by moving $t_i$ to the top of 
 $W_i'$. Consider now the extension $P' = \langle
V_1', \ldots, V_n'\rangle$ of $P=P_X$ where $V_i'$ is obtained from
$W_i'$ by removing all the $t$'s. Since exactly one $t_i$ appears in
the first $K+1$ positions of $W_i'$, the $K$ candidates approved in
$V_i'$ are exactly the $K+1$ candidates approved in $W_i'$ minus
$t_i$. From this we conclude that for every $x \in X$ we have
$S_{K+1}(x,P') = S_{K}(x,R')$ and for every $i = 1,\ldots, k$ we
have $S_{K+1}(y_i,P') = S_{K}(y_i,R')$.
Therefore $x^*$ is a possible cowinner in $P'$ and a possible cowinner in $P$. 
\endproof

\begin{prop}\label{44}
Deciding whether a candidate is a possible cowinner for $K$-approval with respect to the addition of
$k$ new candidates is \npc\ for each $(K,k)$ such that $K \geq 3$ and $k \geq 3$.
\end{prop}
\beginproof
Since deciding whether $x^*$ is a possible cowinner for $3$-approval
with respect to the addition of $3$ new candidates is {\sf NP}-hard, using
inductively the reductions of Lemma \ref{Kappk>3} and Lemma \ref{KappK+1} shows that {\sf NP}-hardness
propagates to every $(K, k) \geq (3, 3)$. Hence, the problem
PcWNC($k$) for $K$-approval is
\npc\ for any fixed pair of values $K \geq 3$ and $k \geq 3$. 
\endproof

We summarize the results obtained in this Section by the following table:

\begin{center}
\begin{tabular}{c|ccc}
& \nicobf{$k=1$} &\nicobf{$k=2$}& \nicobf{$k \geq 3$} \\ \hline
plurality & \p (Prop. \ref{prop-plura}) & \p (Prop. \ref{prop-plura}) & \p (Prop. \ref{prop-plura}) \\
2-approval & \p (Prop. \ref{prop-kapp}) & \p (Coro. \ref{coro-2app}) & \p (Coro. \ref{coro-2app})\\
$K$-approval, $K \geq 3$ & \p (Prop. \ref{prop-kapp}) & \p (Prop. \ref{prop-2cand}) & \npc ~(Prop. \ref{44})
\end{tabular}
\end{center}

Observe that it would also be possible to address the {\sc PcWNC}$(k)$ problem (for $K \geq 3$ and $k \geq 3$) by
working out a direct polynomial reduction from 3-DM, as done in
Proposition \ref{PropNP-completeApproval}. This would however result
in a much less readable proof. One further interest of the proposed
reduction is to show how it is possible to ``neutralize'' the
(extended) power induced by adding more candidates by also adding
one more (dummy) candidate in the initial instance. Intuitively, by
setting the score of dummy candidate $t$ to $2 S_K(x^*,P)$, a single
new candidate $y_i$ will have to be ``consumed'' to ensure that $t$
does not win. More generally, the same proof holds even if $K$ and
$k$ depend on the instance (i.e. are not constant). 
If we allow $f(n)$ new
candidates (where $f$ is polynomially bounded function) instead of
$k$ a constant, the hardness result also holds (in the proof of
Lemma \ref{Kappk>3}, we  duplicate each vote $V$ $f(n)$ times by
adding candidates $z_i$ for $i=1,\dots,f(n)$ instead of $z$ and we
add dummy voters and candidates). Formally, we replace the
construction given in Lemma \ref{Kappk>3} by:

\begin{itemize}
\item the set of candidates is $X' = X \cup \{z_1,\ldots,z_{f(n)}\} \cup \{t_{i,\ell}^1,t_{i,\ell}^2 \ |
\ 1 \leq i \leq  2S_3(x^*,P), 1 \leq \ell \leq f(n) \}$;

\item there are $n+2f(n)S_3(x^*,P)$ votes:
\begin{itemize}
\item for every vote $V_j$ in $P$ we have a vote $W_j$ in $Q$ whose first three candidates are
 the same as in $V_j$ and in the same order, and the other candidates are in an arbitrary order.

\item for every $i = 1, \ldots, 2  S_3(x^*,P)$ and $\ell = 1, \ldots, f(n)$,
we have a vote $U_{i,\ell}$ in which the first 3 candidates
 are $t_{i,\ell}^1, t_{i,\ell}^{2}, z_\ell$, the remaining candidates being ranked arbitrarily.
\end{itemize}
\end{itemize}
Finally, $Y'=\{y_1,\ldots, y_{k+f(n)}\}$ are the new candidates.

Thus, using above construction, Lemma \ref{Kappk>3} and Proposition
\ref{PropNP-completeApproval}, we obtain that for any $\varepsilon
\in (0;1)$,  {\sc PcWNC}$(f(n))$ for 3-approval is  an \np-complete
problem  where
$f(n)=\Theta(|N|^{1-\varepsilon})=\Theta(|X|^{1-\varepsilon})$ (by
setting $f(n)=|N|^r$ in the above construction where $r$ is a
constant arbitrarily large). On the other hand, {\sc PcWNC}$(f(n))$ for
$K$-approval is a problem which can be solved in polynomial time when $f(n)=K \cdot |N|$, i.e., when the
number of new candidates is $K$ times the number of voters. 

Note that some candidates
(other than the new candidates) can be \emph{necessary} cowinners with
$K$-approval. Specifically, each candidate $x_i$ such that
$S_{K-k}(P_X,x_i)=n$ is a necessary cowinner, since she is approved
by all voters and there are not enough new candidates to push her
(in at least one vote) out of the set of approved candidates.

\section{Borda}\label{borda}
Let us now consider the Borda rule ($r_B$). 
Characterizing possible Borda cowinners when adding candidates is easy
due to the following lemma:

\begin{lemma}\label{lemma-borda}
Let $P_{X}$ be an $X$-profile where $X=\{x^*\}\cup
\{x_1,\ldots,x_p\}$ and let $Y=\{y_{1},\ldots, y_{k}\}$ be a set of $k$
new candidates. Let $r_{\vec{s}}$ be a scoring rule for $p+k$ candidates\footnote{In this lemma we do not have to deal with profiles with less than $p+k$ candidates, therefore it is not necessary to mention how $r_s$ is derived for fewer candidates than $p+k$.}
defined by the vector $\vec{s} = \left\langle s_{1}, \ldots , s_{p}, \ldots ,
s_{p+k}\right\rangle $ such that
$(s_{i}-s_{i+1})\le(s_{i+1}-s_{i+2})$ for all $i$. $x^*\in X$ is a
possible cowinner for $P_{X}$ w.r.t. the addition of $k$ new
candidates for the scoring rule  $r_{\vec{s}}(P)$ iff $x^* \in r(P)$
where $P$ is the profile on $X\cup Y $ obtained from $P_{X}$ by
putting $y_{1}, \ldots , y_{k}$ right below $x^*$ (in arbitrary order)
in every vote of $P_{X}$.\end{lemma}
\beginproof
We show that it is never strictly better to put the new candidates
anywhere but right below $x$ in the new profile. 
Let $P$ be an extension of $P_X$ in which $x^*$ is a cowinner, and
assume there is a vote $V\in P$ and a new candidate 
$y$ such that either (i) $y \succ_v x^*$ or (ii) there exists at least one
candidate $x'$ such that $x^* \succ_v x' \succ_v y$.

If we are in case (i), let us move $y$ right below $x^*$; let $V'$
be the resulting vote, and $P'$ the resulting profile. Obviously,
$S_{\vec{s}}(y,P') \leq S_{\vec{s}}(y,P)$ and $S_{\vec{s}}(x^*,P')
\geq S_{\vec{s}}(x^*,P)$, therefore $S_{\vec{s}}(x^*,P') \geq
S_{\vec{s}}(y,P')$. For each candidate $z$ such that $y \succ_v z
\succ_v x^*$, let $i$ be the rank of $z$ in $v$ and $j > i$ be the
rank of $x^*$ in $v$. Then $(S_{\vec{s}}(z,P') -
S_{\vec{s}}(x^*,P')) - (S_{\vec{s}}(z,P)- S_{\vec{s}}(x^*,P)) =
(s_{i-1}-s_{j-1}) - (s_{i}-s_j) = (s_{i-1}-s_i) - (s_{j-1}-s_j) \leq
0$, therefore $S_{\vec{s}}(x^*,P')  \geq S_{\vec{s}}(z,P') $. The
scores of all other candidates are left unchanged, therefore $x^*$ is
still a cowinner in $P'$. By applying this process iteratively for all new candidates and in
all votes until (i) no longer holds, we obtain a profile $Q$
in which $x^*$ is a cowinner, and such that $x^*$ is ranked above all new candidates in every vote. 

Now, if (ii) holds for some new candidate $y$ and some vote $V$ of $Q$, then we move $y$ upwards, 
right below $x^*$; let $V'$ be the resulting vote and $Q'$ the resulting
profile. The score of $y$ improves, but since $y$ is still ranked above all new candidates in every vote of $Q'$, 
we have $S_{\vec{s}}(x^*,Q') \geq S_{\vec{s}}(y,Q')$. For each candidate $z \in X \cup Y \setminus \{x^*,y\}$
such that $x^* \succ_v z \succ_v y$ in vote $V$, $z$ moves down one
position in $Q'$, therefore $S_{\vec{s}}(z,Q') \leq S_{\vec{s}}(x',Q)
\leq S_{\vec{s}}(x^*,Q) =S_{\vec{s}}(x^*,Q')$. The scores of all
other candidates do not change, therefore $x^*$ is still a cowinner
in $Q'$. By applying this process iteratively and in all votes, until
(ii) no longer holds, and we obtain a profile in which
$x^*$ is a cowinner and neither (i) nor (ii) holds. 

We conclude that $x^*$ is a possible cowinner for a profile if and only if
it is a cowinner in an extension of the profile where all new
candidates have been placed right below $x^*$.
\endproof

In words, Lemma \ref{lemma-borda} applies to the rules
where the difference of scores between successive ranks can only
become smaller or remain constant as we come closer to the highest
ranks. This condition is satisfied by Borda (but not by plurality),
by veto, and by rules such as ``lexicographic veto'', where the
scoring vector is $ \langle M^p, M^p-M, M^p - M^2, \ldots, M^p -
M^{p-1}, 0\rangle$ where $M > n$.

The following result then easily follows:

\begin{prop}\label{prop-borda}
Let $P_{X}$ be an $X$-profile where $X=\{x^*\}\cup
\{x_1,\ldots,x_p\}$ and let $Y=\{y_{1},\ldots, y_{k}\}$ be a set of $k$
new candidates. A candidate $x^*$ is a possible cowinner for Borda
with respect to the addition of $k$ new candidates  \iff
$$k \geq \max_{z \in X \setminus \{x^*\}}\frac{S_{B}(z,P_X) - S_{B}(x^*,P_X)}{N_{P_X} (x^*,z)}$$
\end{prop}

\beginproof
By Lemma \ref{lemma-borda}, $x^*$ a possible cowinner if and only if it is a cowinner in the  $X \cup \{y_1 , \dots , y_k\}$-completion $P$ of $P_X$ where $y_1, \dots , y_k$ are placed right below $x^*$, that is, if and only if 
$S_{B}(x^*,P)=S_{B}(x^*,P_{X})+ kn$. Now, for each vote, all candidates in $X \setminus \{x^*\}$
ranked above $x^*$ get $k$ additional points in the extended vote, while those ranked below $x^*$ keep the same score.
Hence, for every $z \in X \setminus \{x^*\}$ we have $S_{B}(z,P)=S_{B}(z,P_{X})+ k(n-N_{P_X}(x^*,z))$, therefore, $x^*$ is a cowinner in $P$ if and only if 
$S_{B}(x^*,P_{X})+ kn \geq S_{B}(z,P_{X})+ k(n-N_{P_X}(x^*,z))$, which is equivalent to 
$k  \geq [S_{B}(z,P_X) - S_{B}(x^*,P_X)]/ N_{P_X}(x^*,z)$. (We recall that $N_{P_X}(x^*,z)$ stands for the number of votes in $P_X$
ranking $x^*$ above $z$). 
\endproof

In words, checking
whether $x^*$ is a possible cowinner boils down to checking, for each
other candidate $z$, whether there are enough votes where $x^*$ is
preferred to $z$ to compensate for the score difference with this
candidate.  
This means that possible cowinners with respect to adding any
number of new candidates can be computed in polynomial time \jl{for
Borda}, and more generally for any rule satisfying the
conditions of Lemma \ref{lemma-borda}. Note that 
computing possible winners for Borda is 
\np-hard \cite{XiaConitzer11}, therefore, the restriction of the problem to 
candidate addition induces a complexity reduction. 

\begin{example}
Take $X = \{a,b,c,d\}$, $n = 4$, and $P_X = \langle bacd, bacd, bacd, dacb\rangle$.  
The Borda scores in $P_X$ are $S_B(a,P_X)=8$, $S_B(b,P_X)=9$, $S_B(c,P_X)=3$, and $S_B(d,P_X)=4$, while $N(a,b)=1$, $N(a,c)=4$, $N(a,d)=3$, $N(b,c)=3$, $N(b,d)=3$,  $N(c,d)=3$, and for all $x,y$, $N(x,y)=4-N(y,x)$. 
Let $\delta(x,z) = S_{B}(z,P_X) - S_{B}(x,P_X)/N_{P_X}(x,z)$. 
The following matrix gives the values of $\delta(x,z)$ for the possible pairs of distinct candidates (for the sake of readability, non-positive values are denoted by $\leq 0$).
%\\
\[
\begin{array}{|c|cccc|c|}
\hline
\delta(x,z) & a & b & c & d & \max \\
\hline
a & - & 1 & \leq 0 &  \leq 0 & 1 \\
b & \leq 0 & - &  \leq 0 & \leq 0 & \leq 0 \\
c & + \infty & 5 & - &\leq  0 & + \infty \\
d & 5 & 6 & 1 & - & 6 \\
\hline
\end{array}
\]
\Omit{
Hence the values of $\delta(x,z)$ for each pair $x,z \in X$:
\begin{itemize}
\item $\delta(b,x) = 0$ for each $x \neq b$;
\item $\delta(a,b) = \frac{9-8}{1} = 1$; $\delta(a,c) = \delta(a,d) = 0$;
\item $\delta(c,a) = \frac{8-4}{0} = +\infty$;  $\delta(c,b) = \frac{9-4}{1} = 5$;  $\delta(c,d) = 0$;
\item $\delta(d,a) = \frac{8-3}{1} =5$; $\delta(d,b) = \frac{9-3}{1} =6$; $\delta(d,c) = \frac{4-3}{1}=1$.
\end{itemize}
}
Applying Proposition \ref{prop-borda}, $b$ is a possible cowinner whatever the value of $k$,  $a$ is a possible cowinner 
if and only if $k \geq 1$, $d$ is a possible cowinner if and only if $k \geq 6$, $c$ is not a possible cowinner whatever the value of $k$\footnote{This is so because $c$ is always ranked below $a$. We make this intuition clear in Section \ref{related}.}. 
Note that for $k \geq 6$, $d$ is a possible cowinner whereas  $c$ is not, although $c$ has a higher Borda score than $d$ in $P_X$.
\end{example}

\section{Hardness with a single new candidate}\label{hardone}

Even though we have seen that the possible cowinner problem can be
\np-hard for some scoring rules, \np-hardness required the addition
of several new candidates. We now show that there exists a scoring
rule for which the possible cowinner problem is \np-hard with respect
to the addition of {\em one} new candidate.

The scoring rule we use is very simple: it allows each voter to
approve exactly 3 candidates, and offers 3 different levels of
approval (assigning respectively 3,2,1 points to the three preferred
candidates). Let $r_\Delta$ be the scoring rule defined by the
vector $\vec{s}=\langle 3,2,1,0,\ldots, 0\rangle$ with $m-3$ 0's
completing the vector.

\begin{prop}\label{PropNP-completeVoting}
Deciding if $x^*$ is a possible cowinner for $r_\Delta$ with
respect to the addition of one candidate is \np-complete.
\end{prop}

\beginproof
This problem is clearly in \np. The hardness proof is quite similar
to that of Proposition \ref{PropNP-completeApproval}. Let
$I=(\mathcal{C},A\times B\times C)$ be an instance of \textsc{3-DM}
with $n'\geq 5$ and $\forall z\in A\cup B\cup C$, $d(z)\in\{2,3\}$.
From $I$, we build an instance $\langle N, X, P_X, 1, x^* \rangle$ 
of the PcWNC problem as follows. The
set $X$ of candidates contains $x^*$, $X_1= \{x'_i,y'_i,z'_i:1\leq
i\leq n'\}$ where $x'_i,y'_i,z'_i$ correspond to elements of $A$,
$B$ and $C$ respectively and a set $X_2$ of dummy candidates. The
set $N$ of voters contains $N_1=\{v^e:e\in\mathcal{C}\}$ and a set
$N_2$ of dummy voters. For each voter $v_i\in N$, we only indicate
the vote for the first three candidates. So, the vote
$V_i=(t_1,t_2,t_3)$ means that candidate $t_i$ receives $4-i$
points. The vote $V_e$ of voter $v^e$ is $(x'_i,y'_j,z'_k)$ where
$e=(a_i,b_j,c_k)\in\mathcal{C}$. The preferences of dummy voters are
such that \jm {$(a)$ the score of the candidates in $X$ satisfies
$\forall x\in X_1$, $S_{\vec{s}}(x,P_X)=3n'+1$,
$S_{\vec{s}}(x^*,P_X)=3n'$ and $\forall x\in X_2$,
$S_{\vec{s}}(x,P_X)\leq 3$ and $(b)$ each voter in $N_2$ ranks at
most one candidate of $\{x'_i,y'_i,z'_i:1\leq i\leq n'\}$ in the
first three positions, and if he ranks one in second position, then
$x^*$ occurs in third position.}

Formally, the instance of the PcWNC problem is built as follows.
The set of voters is $N=N_1\cup N_2$ where
$N_1=\{v^e:e\in\mathcal{C}\}$ and $N_2=N_{A}\cup
N_{B}\cup N_{C}\cup N_{x^*}$, the set of
candidates is $X=X_1\cup X_2\cup \{x^*\}$ where $X_1= X'\cup Y'\cup
Z'$ with $X'=\{x'_1,\dots,x'_{n'}\}$, $Y'=\{y'_1,\dots,y'_{n'}\}$,
$Z'=\{z'_1,\dots,z'_{n'}\}$ and $X_2=X_{A}\cup
X_{B}\cup X_{C}\cup X_{x^*}$. These %two 
sets are defined as follows:\\

\begin{itemize}
\item[$\bullet$] $X_{A}=\{x_i^j:1\leq i\leq n'$,
$1\leq j\leq 2\left(n'-d(a_i)\right)\}$.

\item[$\bullet$] $X_{B}=\{y_i^j:1\leq i\leq n'$,
$1\leq j\leq 2\left(3n'-2d(b_i)+1\right)\}$.

\item[$\bullet$] $X_{C}=\{z_i^j:1\leq i\leq n'$,
$1\leq j\leq 2\left(3n'-d(c_i)+1\right)\}$.

\item[$\bullet$] $X_{x^*}=\{x^*_i:i=1\leq i\leq 2n'\}$.
\end{itemize}

The set of all $X$-votes $\P_X$ is given by:
\begin{itemize}
\item[$\bullet$] $N_{A}=\{v_{i,j}^{A}:1\leq i\leq n'$,
$0\leq j \leq \left(n'-d(a_i)-2\right)\}\cup
\{v_{i}^{A,j}:1\leq i\leq n'$, $j=1,2\}$. The vote
$V_{i,j}^{A}$ of $v_{i,j}^{A}$ is
$V_{i,j}^{A}=(x'_i,x_i^{2j+1},x_i^{2j+2})$. Note that
$n'-d(a_i)-2\geq 0$. The vote of  $v_{i}^{A,j}$ is
$V_{i}^{A,j}=(x_i^{2\left(n'-d(a_i)-1\right)+j},x'_i,x^*)$.

\item[$\bullet$] $N_{B}=\{v_{i,j}^{B}:1\leq i\leq n'$,
$0\leq j \leq 3n'-2d(b_i)\}$. The vote of $v_{i,j}^{B}$
is $V_{i,j}^{B}=(y_i^{2j+1},y_i^{2j+2},y'_i)$.

\item[$\bullet$] $N_{C}=\{v_{i,j}^{C}:1\leq i\leq n'$,
$0\leq j \leq 3n'-d(c_i)\}$. The vote of $v_{i,j}^{C}$
is $V_{i,j}^{C}=(z_i^{2j+1},z_i^{2j+2},z'_i)$.

\item[$\bullet$] $N_{x^*}=\{v_j^{x^*}:0\leq j \leq n'-1\}$. The vote of
$v_j^{x^*}$ is $V_j^{x^*}=(v_{2j+1}^{x^*},v_{2j+2}^{x^*},x^*)$. Note
that $n'-1\geq 0$.

\item[$\bullet$] $N_1=\{v^e:e\in\mathcal{C}\}$. The vote of
$v^e$ is $V_e=(x'_i,y'_j,z'_k)$ where
$e=(a_i,b_j,c_k)\in\mathcal{C}$.
\end{itemize}

We claim that $I$ admits a perfect matching $M\subseteq \mathcal{C}$
if and only if $x^*$ becomes a possible cowinner by adding a new
candidate $y_1$.
Observe that the scores of the candidates in $X$ satisfy:
\begin{itemize}
\item[$(i)$] $\forall x\in X_1$, $S_{\vec{s}}(x,P_X)=3n'+1$.
\item[$(ii)$]  $S_{\vec{s}}(x^*,P_X)=3n'$.
\item[$(iii)$] $\forall x\in X_2$, $S_{\vec{s}}(x,P_X)\leq 3$.
\end{itemize}

\jm{Items $(i)$, $(ii)$ and $(iii)$ correspond to the conditions
$(a)$ and $(b)$ described previously.} 
For instance, each candidate $x'_i$ from $X_1$ gets respectively 3, 3, and 2 points from the votes $V_e$, $V_{i,j}^A$, and $V^{A,j}_i$, summing up to $3 d(a_i) + 3(n' -d(a_i) -1) +2 = 3n'+1$. The reader can easily check that the conditions also hold for all other candidates. 

Let $y_1$ be the new candidate. By construction of this scoring
rule, we must decrease the score of candidates in $X$ which dominate
the score of $x^*$, that is the
candidates of $X_1$ using $(i)$ and $(iii)$.\\

%%%%%%
Let $P'$ be a $X\cup \{y_1\}$-profile such that $x^*$ is a cowinner.
Let us focus on candidates in $X'$. In order to reduce
the score of $x'_i$ by 1, we must modify the preference  for at least one
voter $v^e$ or $v_{i,j}^{A}$ or $v_{i}^{A,j}$. If we modify it for
some voter in $v_{i}^{A,j}$, then the score of $x'_i$ (with respect
to $v_{i}^{A,j})$ decreases by one if and only if the score of $x^*$ (with
respect to $v_{i}^{A,j})$ also decreases  by one. In conclusion, we
must modify the preference of $x'_i$ for at least one voter $v^e$ or
$v_{i,j}^{A}$. By construction, each such voter must put $y_1$ in
top position and then, the score of $y_1$ increases by 3 at each
time. Since there are $n'$ candidates in $X'$, we deduce
$S_{\vec{s'}}(y_1,P')\geq 3n'$; From above remark, we also get
$S_{\vec{s'}}(x^*,P')\leq S_{\vec{s'}}(x^*,P_X)=3n'$. Thus for each
$i\in \{1,\dots,n'\}$, exactly one voter among those of $v^e$ or
$v_{i,j}^{A}$ must put candidate $y_1$ in top position. Finally, if
it is one voter $v_{i,j}^{A}$, then we deduce
$S_{\vec{s'}}(y_1,P')>3n$ because the score of $Y'\cup Z'$ must also
decrease, which is not possible since $y_1$ will then win.

Following a line of reasoning similar to the one developed in the
proof of Proposition \ref{PropNP-completeApproval}, we conclude that
for each $i\in \{1,\dots,n'\}$, exactly one voter among those of
$v^e$ must put candidate $y_1$ in top position. Since the score
of $Y'\cup Z'$ must also decrease by 1, we deduce that $x^*$ is a
possible cowinner if and only if $M=\{e\in \mathcal{C}:$ $y_1$ is in top position
in vote $V_e\}$ is a perfect matching of $\mathcal{C}$ (for the
remaining voters, $y_1$ is put in last position).
\endproof

This rule shows that it may be difficult to identify
possible cowinners with a single additional candidate. Giving a
characterization of all rules possessing this property is an open
problem.

\section{Related work}\label{related}

\subsection{The possible winner problem}\label{pwp}

The possible winner problem was introduced in \cite{KonczakLang05}: given an incomplete profile $P = \langle V_1, \ldots, V_n\rangle$ where each $V_i$ is a partial order over the set of candidates $X$, $x$ is a possible winner for $P$ given a voting rule $r$ if there exists a complete extension $P' = \langle V_1', \ldots, V_n'\rangle$ of $P$, where each $V_i'$ is a linear order on $X$ extending $V_i$, such that $r(P') = x$. Possible winners are defined in a similar way for a voting correspondence $C$, in which case we say that $x$ is a possible {\em cowinner} if there exists an extension $P'$ of $P$ such that $x \in C(P')$. Clearly, the possible winner problem defined in this paper is a restriction of the general possible winner problem to the following set of incomplete profiles:
\begin{center}
(Restr) {\em there exists $X' \subseteq X$ such that for every $i$, $V_i$ is a linear order on $X'$}
\end{center}
As an immediate corollary, the complexity of the possible (co)winner problem with respect to candidate addition is at most as difficult as that of the general problem. This raises the question whether (Restr) leads to a complexity reduction for the scoring rules we have considered here.

The possible (co)winner problem for scoring rules has received a significant amount of attention in the last years. Xia and Conitzer \cite{XiaConitzer11} proved that the problem was \np-complete for the Borda rule, and more generally for scoring rules whose scoring vector contains four consecutive, equally decreasing values, followed by another strictly decreasing value. Betzler and Dorn \cite{BetzlerDorn09} went further by showing that {\sf NP}-completeness holds more generally for all pure\footnote{A (family of) scoring rules $(r_m)_{m \geq 1}$ is  pure if for each $m$, the scoring vector for $m+1$ candidates is  obtained from the scoring vector for $m$ candidates by inserting an additional score at an arbitrary position. All interesting families of scoring rules are pure; this is in particular the case for $K$-approval and Borda.} scoring rules, except plurality, veto, and scoring rules whose vector $s^m$ is of the form $s^m = \langle 2, 1,\ldots, 1, 0 \rangle$ for large enough values of $m$. The issue was finally closed by Baumeister and Rothe \cite{BaumeisterRotheECAI10}, who showed that the problem for $s^m = \langle 2, 1,\ldots, 1, 0 \rangle$ is {\sf NP}-complete as well. These results compare to ours in the following way: all our {\sf NP}-hardness results strengthen the known {\sf NP}-hardness results for the general possible winner problem, while our polynomiality results show a complexity reduction induced by (Restr). 

Two recent papers give results about the {\sc PcWNC} problem for other voting rules. Xia {\em et al.} \cite{XLM11} give results about the possible (co)winner with new \nicobf{candidates} for other voting rules: they showed that {\sc PWNC} and {\sc PcWNC} are \np-complete for Bucklin and maximin, that {\sc PcWNC} is \np-complete for Copeland$_0$, and they give several results for approval voting, depending on how the extension a vote is defined. Baumeister {\em et al.} \cite{BaumeisterRR11} generalize our Proposition \ref{PropNP-completeVoting} by showing that the {\sc PcWNC} problem is \np-complete for any pure scoring rule of the form $\langle \alpha_1, \alpha_2, 1; 0, \ldots, 0\rangle$; they also give \np-completeness results for plurality and 2-approval when voters are weighted.

Results about the {\sc PcWNC} known so far (except our Proposition \ref{PropNP-completeVoting}  and its generalization by \cite{BaumeisterRR11}) are summarized in the following table. For the sake of completeness, we also mention the complexity of the other prominent subproblem of the possible cowinner problem, namely unweighted coalitional manipulation. 

\begin{center}
{\small
\begin{tabular}{l|l|l|l|}
& general problem & candidate addition & manipulation\\ \hline
plurality and veto &  \p & \p (Prop. \ref{prop-plura})
& \p \\
Borda &  \npc\ \cite{XiaConitzer11} & \p (Prop. \ref{prop-borda})& \npc\ \cite{BetzlerNiedermeierWoegingerIJCAI11,DaviesEtAlAAAI11} \\
2-approval &  \npc\ \cite{BetzlerDorn09} & \p \jl{(Coro. \ref{coro-2app})} & \p  \\
$K$-approval, $K \geq 3$ & \npc\ \cite{BetzlerDorn09} &  \npc\ (Prop. \ref{PropNP-completeApproval}) & \p\\
Bucklin & \npc\ \cite{XiaConitzer11} & \npc\ \cite{XLM11} &   \p\ \cite{XZPCR09}\\
maximin & \npc\ \cite{XiaConitzer11} & \npc\ \cite{XLM11} &   \npc\ \cite{XZPCR09}\\
Copeland$_0$ & \npc\ \cite{XiaConitzer11} & \npc\ \cite{XLM11} &   \npc\ \cite{FHS08}\\
\end{tabular}
}
\end{center}

Another interesting line of work is the {\em parameterized} complexity of the possible winner problem for scoring rules, which has been investigated in \cite{BetzlerHemmannNiedermeier09}. Among other results, they show that for all scoring rules, the problem is fixed-parameter tractable with respect to the number of candidates (in particular, when the number of candidates is bounded by a constant, the problem becomes polynomial-time solvable). This polynomiality result clearly \nicobf{holds in} the possible winner problem with respect to candidate addition, with some caution: the number of candidates here is the {\em total} number of candidates (the initial ones plus the new ones); this result has practical impact in some situations mentioned in the introduction, such as finding a date for a meeting, where the number of candidates is typically low. 

We end this subsection by mentioning other works on the possible winner problem and its variants and subproblems, that are less directly connected to our results. The possible winner problem has also been studied from the probabilistic point of view by Bachrach {\em et al.} \cite{BachrachEtAlAAAI10}, where the aim is to count the number of extensions in which a given candidate is the winner. 
Such a probabilistic analysis is highly relevant in candidate-adding situations: given $P_X$, a number $k$ of new candidates, and a prior probability distribution on votes, computing the probability that a given candidate $x \in X$ will be the winner, or that one of the initial (resp. new) candidates will be the winner, is extremely interesting.\footnote{Note that if the voting rule is \nicobf{insensitive to the identity of candidates ({\em i.e.} neutral)}, then although the prior probability that one of the $k$ new candidates will be a cowinner under the impartial culture assumption is at least $\frac{k}{|X|+k}$, this is no longer the case when $P_X$ is known: for instance, let us use plurality and consider the profile $P_X = \langle ab, ab, ab\rangle$, and let the number of new candidates be one. For a third candidate to be a cowinner, he either needs to be placed first in all three votes (which occurs with probability $\frac{1}{27}$), or to be placed first in two votes, but not in the third vote (which occurs with probability $\frac{6}{27}$); therefore the probability that the new candidate is a cowinner in the completed profile is only $\frac{7}{27}$.}

\subsection{Control via adding candidates}\label{control}

The possible winners with respect to the addition of candidates is
highly reminiscent of constructive control by the chair via adding
candidates --- this problem first appeared in
\cite{BartholdiToveyTrick92} and was later studied in more depth
for many voting rules, see {\em e.g.},
\cite{HemaspaandraHR07,FHH09}. However, even if a voting situation
where new candidates are added looks similar to an instance of
constructive control  by adding candidates, these
problems differ significantly. In control via adding candidates, the
input consists of a set of candidates $X$, a set of ``spoiler''
candidates $Y$, and a full profile $P_{X \cup Y}$:  the chair knows
how the voters would vote on the new candidates; the problem is to
determine whether a given candidate $x^*$ can be made a winner by
adding at most $k \leq |Y|$ candidates from $Y$. In the possible
winner problem  with respect to candidate addition, we have to take
into account all possible ways for voters to rank the new
candidates.  In spite of their significant differences,
there is a straightforward connection between these problems: if an
instance $\langle N,X, P_{X \cup Y}, x^*, k\rangle$ of control via
adding candidates is positive, then $x^*$ is a possible winner in
$P_X$ with respect to the addition of $k$ new candidates (the voting
rule being the same in both problems).

Bartholdi {\em et al.} \cite{BartholdiToveyTrick92} noted that a voting
rule is immune to control by adding candidates as soon as it
satisfies the Weak Axiom of Revealed Preference (WARP), which
requires that the winner among a set of candidates $W$ to be the winner
among every subset of candidates to which he belongs \cite{Plott76};
formally: 
 {\em for any $Z \subseteq W$, if $r(P_W) \in Z$ then $r(P_Z) = r(P_W)$.}
This property can be used in a similar way for the possible winner problem with respect to candidate addition. Obviously, if the voting rule $r$ satisfies WARP,  then any possible winner from $X$ is a winner for the current profile $P_X$\footnote{In order for the converse to hold, we must add one more condition, such as {\em consensus} (a Pareto-dominated candidate cannot be elected). Then, if the winner for the current profile $P_X$ is $x$, by ranking all new candidates at the bottom of all votes, none of them can be the winner in $P_{X \cup Y}$, and by WARP, no candidate $x' \in X \setminus \{x\}$ can either, therefore $x$ is a possible winner for $P_X$ with respect to candidate addition.}. 
Unfortunately, this social-choice theoretic property is very strong: \cite{DuttaJacksonLebreton01} show that a voting rule satisfies this property (there, it is called {\em candidate stability}) and unanimity if and only if it is dictatorial.

\subsection{Cloning}\label{cloning}

Finally, the possible winner problem via candidate addition is closely related to manipulation by candidate cloning. Independence of clones was first studied in \cite{TidemanSCW1987}, further studied in \cite{LLL96,LaslierSCW2000}, and a variant of this property was recently considered from the computational point of view in  \cite{ElkindEtAlAAAI2010}. The main difference between $x$ being a possible winner with respect to candidate addition and the existence of a candidate cloning strategy so that $x$ or one of its clones becomes the winner, as in \cite{ElkindEtAlAAAI2010}, is that candidate cloning requires a candidate and its clones to be contiguous in all votes. In other terms, whereas our problem considers the introduction of genuinely new candidates, 
cloning merely introduces copies of existing ones.

The complexity of this problem is considered by Elkind {\em et al.} \cite{ElkindEtAlAAAI2010} for several voting rules.
Although the proposed model allows for the possibility of having a bounded number of new clones (via a notion of cost),
most of their results focus on the case of unboundedly many clones. Therefore, to be able to compare their results with ours,
we should first say something about the variant of the possible winner problem with respect to candidate addition,
{\em when the number of new candidates is not known beforehand and can be arbitrarily large}. The definitions of voting
situations and possible winners are straightforward adaptations of Definitions \ref{situation} and \ref{posswin}:
a voting situation is now a triple $\Sigma = \langle N, X, P_X \rangle$ and  $x^*$ is a possible cowinner with respect to $\Sigma$
and $r$ if there exists an integer $k$ and a set $Y$ of cardinality $k$ such that there is a $(X \cup Y)$-profile $P$ extending $P_X$
such that $x^* \in r(P)$. 
We now give a necessary and sufficient condition for a candidate to be a possible winner, for a class of scoring rules including the Borda rule.

\begin{prop}
\label{pro:unbounded_undom} Let $\mathcal{S}$ be a %normalized
collection of scoring vectors $(s^m), m \geq 1$, such that
\begin{itemize}
\item for every $p$, $(s^m_j), 1 \leq j \leq m$ is strictly decreasing;
\item for all $j,j'\in\mathbb{N}$, (1) $\lim_{m\rightarrow\infty}\frac{s_{j}^{m}-s_{j'}^{m}}{s_1}=0$ and
(2) $\lim_{m\rightarrow\infty}\frac{s_{j}^{m}-s_{m-j'}^{m}}{s_1}=1$.
\end{itemize}
Then, $x^*$ is a possible winner w.r.t. $\langle N,X,P_X, + \infty \rangle$ 
if and only if it is undominated\footnote{We recall that candidate $x$ dominates  candidate
$x'$ if every voter ranks $x$ above $x'$, and that a candidate is undominated if no other candidate dominates it.} in $P_X$. 
\end{prop}

\beginproof
First, suppose $x^*$ is undominated in $P_X$. For any candidate $x_{i}\neq x^*$,
define $\Delta^v_{i}$ as the difference between the score of $x^*$ and
the score of $x_{i}$, divided by $s_1$, in the vote $v$. As in the construction
of Lemma 1, put $k$ new candidates right below $x^*$ in every vote, and let $P'$ be the resulting profile. 
As the value of $k$ grows, for any vote $v$ ranking candidate $x^*$ below
$x_{i}$, the value of $\Delta^v_{i}$ will tend towards $0$ (by condition
1). Also, condition 2 ensures that for each vote $v$ ranking $x^*$ above
$x_{i}$, the value of $\Delta^v_{i}$ tends towards $1$.
Because $x^*$ is undominated, such votes always exist for every candidate $x_{i}\neq x^*$.
Therefore, when $k$ grows, $\sum_{v \in P'} \Delta^v_i$ tends towards the number of votes
ranking $x^*$ above $x_{i}$, which is at least 1. This implies that the score of
$x^*$ will be eventually larger than the score of $x_i$, and this is true for every $x_i \neq x^*$,
therefore 
$x^*$ will eventually become the winner as $k$ grows. Conversely, suppose $x^*$ is dominated
by some candidate $x_{i}$. Because the scores $(s^m_j), 1 \leq j \leq m$ are strictly decreasing,
the score of $x$ will always remain strictly below the score of $x_i$ in the completion of the profile,
hence $x^*$ is not a possible cowinner.
\endproof

Clearly, this large class of  voting rules includes Borda, since it satisfies the conditions of Proposition \ref{pro:unbounded_undom}. 
However, it does not include plurality, and more generally $K$-approval, which violate condition (1).
%, nor the veto  rule which violates condition 2.
Still, a very simple condition can be stated for $K$-approval: a candidate is a possible winner as soon as it is approved at least once. 

\begin{prop} When $r$ is $K$-approval, $x^*$ is a possible winner w.r.t. the addition of  an unbounded number of new candidates \iff  $S_K(x^*,P_X) \geq 1$.
\end{prop}
\beginproof
The condition is obviously necessary. Suppose the condition holds on
a given profile. We extend this profile by taking a set of new
candidates $y_{ij}$ where $1\leq i \leq n$ and $1 \leq j \leq K$.
Consider the $i$-th vote: if $x^*$ is ranked in one the top $k$
positions, put all new candidates at the bottom of the vote.
Otherwise, introduce the new candidates $\{y_{i1}, \dots, y_{iK}\}$
at the top of the vote, and all other new candidates at the bottom.
The score of the new candidates is at most 1, while that of $x_i
\neq x^*$ is at most that of $x^*$ (which is unchanged).
\endproof

Note that for $K \geq 2$ this condition does \emph{not} imply that the candidate is undominated (nor vice-versa).
It does obviously when $K=1$, \emph{i.e.}, for plurality. \\

Let us see now how the above results relate to those in \cite{ElkindEtAlAAAI2010}. We first note that in the case of the Borda rule we have the same condition. Indeed one sees intuitively that Lemma \ref{lemma-borda} tells us that for some voting rules (including Borda), introducing new candidates in a contiguous manner, as with cloning, is the best thing to do. For plurality, again the condition is similar in both cases. However, for $K$-approval as soon as $K>1$, 
the problem becomes hard in the cloning setting whereas it is easy in our setting with an unbounded number of new candidates.

\section{Conclusion}\label{conclu}

In this paper we have considered voting situations where new candidates may show up during the process. This problem increasingly occurs in our societies, as many votes now take place online (through dedicated platforms, or simply by email exchange) during an extended period of time.

We have identified the computational complexity of computing possible winners for some scoring rules. 
Some of them allow polynomial algorithms for the problem (\emph{e.g.} plurality, 2-approval, Borda, veto) regardless of the (fixed) number of new candidates showing up.
For the rules of the $K$-approval family, when $K\geq 3$, the problem remains polynomial only if the number of new candidates is at most $2$.
Finally, we have exhibited a simple rule
where the problem is hard for a single new candidate.

The results address the problem of making some designated candidate a cowinner, which is similar to $x$ being unique
winner under the assumption of the most favourable tie-breaking. In the other extreme case (if we want $x$ to be a strict winner, \emph{i.e.}, to win regardless of the tie-breaking rule), the results are easily adapted: for instance, the inequalities in Proposition \ref{prop-plura} and \ref{prop-borda} become strict. For $K$-approval, the first condition of Proposition \ref{prop-kapp} becomes strict but the second one should now read \begin{small}$S_K(P_X,x) \geq \sum_{x_i \in X}\max(0, S_K(P_X, x_i) - S_K(P_X,x)+1)$\end{small}. As for veto, all other initial candidates need to be vetoed at least once. The hardness proofs can also be readily adapted to the unique winner setting.
A more general treatment would require cumbersome expressions, and is also somewhat problematic since the identities of the new candidates are not known anyway (making it difficult to specify easily a tie-breaking rule on these candidates).

As for future work, a first direction to follow would be to try to obtain more general results for scoring rules, as those obtained by \nicobf{Betzler and Dorn} \cite{BetzlerDorn09} for the general version of the possible winner problem. Extending the study to other families of voting rules, such as rules based on the majority graph, is also worth investigating. 

Of course, identifying possible winners is not the end of the story. In practice, as mentioned earlier, one may for instance also be interested in a refinement of this notion: knowing how likely it is that a given candidate will win. % is often a very valuable information.
Another interesting issue consists in designing elicitation protocols when the preferences about the `old' candidates are already known. In this case, a trade-off occurs between the storage cost and communication cost, since keeping track of more information is likely to help reduce the burden of elicitation.

\paragraph{Acknowledgements.}We are very much indebted to the reviewers of previous versions of this paper for their extremely detailed and relevant comments.

\bibliographystyle{plain}

%\bibliography{newcandidates}

\begin{thebibliography}{10}

\bibitem{BachrachEtAlAAAI10}
Y.~Bachrach, N.~Betzler, and P.~Faliszewski.
\newblock Probabilistic possible-winner determination.
\newblock In {\em Proceedings of AAAI-10},  pages 697--702, 2010.

\bibitem{BartholdiToveyTrick92}
J.~Bartholdi, C.~Tovey, and M.~Trick.
\newblock How hard is it to control an election?
\newblock {\em Social Choice and Welfare}, 16(8-9):27--40, 1992.

\bibitem{BaumeisterRotheECAI10}
D.~Baumeister and J.~Rothe.
\newblock Taking the final step to a full dichotomy of the possible winner
  problem in pure scoring rules.
\newblock In {\em Proceedings of ECAI 2010}, pages 1019--1020, 2010.

\bibitem{BaumeisterRR11}
D.~Baumeister, M. Roos and J.~Rothe.
\newblock Computational complexity of two variants of the possible winner problem.
\newblock {\em Proceedings of AAMAS-11}, pages 853--860, 2011.


\bibitem{BetzlerDorn09}
N.~Betzler and B.~Dorn.
\newblock Towards a dichotomy of finding possible winners in elections based on
  scoring rules.
\newblock In {\em Proceedings of MFCS 2009}, volume 5734 of {\em Lecture Notes in
  Computer Science}, pages 124--136. Springer, 2009.

\bibitem{BetzlerHemmannNiedermeier09}
N.~Betzler, S.~Hemmann, and R.~Niedermeier.
\newblock A multivariate complexity analysis of determining possible winners
  given incomplete votes.
\newblock In {\em Proceedings of IJCAI-09}, pages 53--58, 2009.

\bibitem{BetzlerNiedermeierWoegingerIJCAI11}
N. Betzler, R. Niedermeier and G. Woeginger. 
\newblock Unweighted Coalitional Manipulation Under the Borda Rule is NP-Hard
\newblock In {\em Proceedings of IJCAI-11}, 55-60, 2011.

\bibitem{CLMR09}
Y.~Chevaleyre, J.~Lang, N.~Maudet, and G.~Ravilly-Abadie.
\newblock Compiling the votes of a subelectorate.
\newblock In {\em Proceedings of IJCAI-09}, pages 97--102, 2009.

\bibitem{ConitzerSandholm02a}
V.~Conitzer and T.~Sandholm.
\newblock Complexity of manipulating elections with few candidates.
\newblock In {\em Proceedings of AAAI-02}, pages 314--319, 2002.

\bibitem{ConitzerSandholm02b}
V.~Conitzer and T.~Sandholm.
\newblock Vote elicitation: complexity and strategy-proofness.
\newblock In {\em Proceedings of AAAI-02}, pages 392--397, 2002.


\bibitem{ConitzerSandholmLang07}
V. Conitzer, T. Sandholm and J. Lang.
\newblock When are elections with few candidates hard to manipulate?
\newblock Journal of the ACM, 54 (3), 1-33, 2007.


\bibitem{DaviesEtAlAAAI11}
J. Davies, G. Katsirelos, N. Narodytska and T. Walsh.
\newblock Complexity of and Algorithms for Borda Manipulation
\newblock In {\em Proceedings of AAAI-11}, pages 657--662, 2011.


\bibitem{DuttaJacksonLebreton01}
B. Dutta and M. Jackson and M. Le Breton.
\newblock Strategic candidacy and voting procedures.
\newblock {\em Econometrica}, Vol. 69, No. 4, pages 1013--1037, 2001.

\bibitem{ElkindEtAlAAAI2010}
E.~Elkind, P.~Faliszewski, and A.~Slinko.
\newblock Cloning in elections.
\newblock In {\em Proceedings of AAAI-10}, pages 768--773, 2010.

\bibitem{ElkindFaliszewskiSlinko09}
E.~Elkind, P.~Faliszewski, and A.~M. Slinko.
\newblock Swap bribery.
\newblock In Proceedings of {\em SAGT}, pages 299--310, 2009.

\bibitem{FHS08}
P. Faliszewski, E. Hemaspaandra and H. Schnoor.
\newblock Copeland voting: ties matter. 
\newblock {\em AAMAS-08} (2), pages 983-990, 2008

\bibitem{FHH09}
P.~Faliszewski, E.~Hemaspaandra, and L.~Hemaspaandra.
\newblock Multimode control attacks on elections.
\newblock In {\em Proceedings of IJCAI-09}, pages 128--133, 2009.

\bibitem{FaliszewskiProcaccia10}
P. Faliszewski and A. Procaccia.
\newblock The AI war on manipulation: are we winning?
\newblock AI Magazine 31(4): 53-64 (2010).

\bibitem{GJ79}
M.~Garey and D.~Johnson.
\newblock {\em Computers and intractability. A guide to the theory of
  {NP}-completeness}.
\newblock Freeman, 1979.

\bibitem{HAKW09}
N.~Hazon, Y.~Aumann, S.~Kraus, and M.~Wooldridge.
\newblock Evaluation of election outcomes under uncertainty.
\newblock In {\em Proceedings of AAMAS-08}, pages 959--966, 2009.

\bibitem{HemaspaandraHR07}
E.~Hemaspaandra, L.~Hemaspaandra, and J.~Rothe.
\newblock Anyone but him: The complexity of precluding an alternative.
\newblock {\em Artificial Intelligence}, 171(5-6):255--285, 2007.

\bibitem{KonczakLang05}
K.~Konczak and J.~Lang.
\newblock Voting procedures with incomplete preferences.
\newblock In {\em Proceedings of the IJCAI-05 Multidisciplinary Workshop on Advances in
  Preference Handling}, 2005.

\bibitem{LaslierSCW2000}
J.-F. Laslier.
\newblock Aggregation of preferences with a variable set of alternatives.
\newblock {\em Social Choice and Welfare}, 17(2):269--282, 2000.

\bibitem{LLL96}
G. Laffond, J. Lain{\'e} and J.-F. Laslier.
\newblock Composition-consistent tournament solutions and social choice functions.
\newblock {\em Social Choice and Welfare}, 13(1):75--93, 1996.

\bibitem{LuBoutilier10}
T. Lu and C. Boutilier.
\newblock The Unavailable Candidate Model. A Decision-Theoretic View of Social Choice.
\newblock {\em  ACM Conference on Electronic Commerce 2010}, pages 263�274, 2010. 

\bibitem{PiniRVW07}
M.S. Pini, F.~Rossi, K.~Brent Venable, and T.~Walsh.
\newblock Incompleteness and incomparability in preference aggregation.
\newblock In {\em Proceedings of IJCAI'07}, pages 1464--1469, 2007.

\bibitem{Plott76}
C.~R. Plott.
\newblock Axiomatic social choice theory: an overview and interpretation.
\newblock {\em American Journal of Political Science}, 20:511--596, 1976.

\bibitem{TidemanSCW1987}
T.~Tideman.
\newblock Independence of clones as a criterion for voting rules.
\newblock {\em Social Choice and Welfare}, 4(3):185--206, 1987.

\bibitem{Walsh08}
T.~Walsh.
\newblock Complexity of terminating preference elicitation.
\newblock In {\em Proceedings of AAMAS-08}, pages 967--974, 2008.

%\bibitem{XiaConitzer08}
%L.~Xia and V.~Conitzer.
%\newblock Determining possible and necessary winners under common voting rules  given partial orders.
%\newblock In {\em Proceedings of AAAI-08}, pages 196--201, 2008.

\bibitem{XiaConitzer11}
L.~Xia and V.~Conitzer.
\newblock Determining Possible and Necessary Winners Given Partial Orders. 
\newblock J. Artif. Intell. Res. (JAIR) 41: 25-67, 2011.


\bibitem{XiaConitzerAAAI10}
L.~Xia and V.~Conitzer.
\newblock Compilation complexity of common voting rules.
\newblock In {\em Proceedings of AAAI-10}, pages 915--920, 2010.

\bibitem{XZPCR09}
L.~Xia, M.~Zuckerman, A.~Procaccia, V.~Conitzer, and J.~Rosenschein.
\newblock Complexity of unweighted coalitional manipulation under some common
  voting rules.
\newblock In {\em Proceedings of IJCAI-09}, pages 348--353, 2009.

\bibitem{XLM11}
\newblock L.~Xia, J. Lang and J. Monnot.
\newblock Possible winners hen new alternatives join: new results coming up!
\newblock In {\em Proceedings of AAMAS-11}, pages 829--836, 2011.

\end{thebibliography}






% \end{thebibliography}

\end{document}